\documentclass[10pt,journal,compsoc]{IEEEtran}

\ifCLASSOPTIONcompsoc
  \usepackage[nocompress]{cite}
\else
  \usepackage{cite}
\fi
\ifCLASSINFOpdf
\else
\fi

\usepackage{times}
\usepackage{epsfig}
\usepackage{graphicx}
\usepackage{amsmath}
\usepackage{amssymb}

\usepackage{xspace}

\usepackage{hyperref}
\hypersetup{colorlinks,linkcolor={blue},citecolor={blue},urlcolor={red}} 
\usepackage{cite}
\usepackage{footnote}
\usepackage{algorithm}
\usepackage{algorithmic}
\usepackage{booktabs} 
\usepackage{bm}
\usepackage{mathtools}
\usepackage{makecell}
\usepackage{enumitem}
\usepackage{caption}
\usepackage{lipsum}
\usepackage{cuted}
\usepackage{subfig}

\usepackage{color,soul}

\makeatletter
\DeclareRobustCommand\onedot{\futurelet\@let@token\@onedot}
\newcommand{\@onedot}{\ifx\@let@token.\else.\null\fi\xspace}
\newcommand{\eg}{\emph{e.g}\onedot} \newcommand{\Eg}{\emph{E.g}\onedot}
\newcommand{\ie}{\emph{i.e}\onedot} \newcommand{\Ie}{\emph{I.e}\onedot}
 
\newcommand{\etc}{\emph{etc}\onedot} \newcommand{\vs}{\emph{vs}\onedot}
 
\newcommand{\etal}{\emph{et al}\onedot}
\makeatother

\begin{document}
%
\title{Manipulating Identical Filter Redundancy for Efficient Pruning on Deep and Complicated CNN}
%
%
%
%

\author{Xiaohan Ding, Tianxiang Hao, Jungong Han, Yuchen Guo, Guiguang Ding
\IEEEcompsocitemizethanks{
	\IEEEcompsocthanksitem Xiaohan Ding, Tianxiang Hao, Guiguang Ding are with the School of Software, Tsinghua University, Beijing 100084, China.
	\IEEEcompsocthanksitem Jungong Han is with the Computer Science Department, Aberystwyth University, SY23 3FL, UK.
	\IEEEcompsocthanksitem Yuchen Guo is with the the Department of Automation, Tsinghua University, Beijing 100084, China.
	\IEEEcompsocthanksitem dxh17@mails.tsinghua.edu.cn\quad beyondhtx, jungonghan77, yuchen.w.guo\protect\\@gmail.com\quad dinggg@tsinghua.edu.cn,.
	\IEEEcompsocthanksitem This work was supported by the National Key R\&D Program of China (No. 2018YFC0806900), the National Natural Science Foundation of China (No. 61571269, 61327902), and the National Postdoctoral Program for Innovative Talents (No. BX20180172).
	\IEEEcompsocthanksitem Corresponding author: Yuchen Guo, Guiguang Ding.}}

\markboth{WORK IN PROGRESS}%
{Ding \MakeLowercase{\textit{et al.}}: Identical is Ideal: Manipulating Filter Redundancy for Efficient Pruning on Networks with Very Deep and Complicated Architectures}
%



\IEEEtitleabstractindextext{%
\begin{abstract}

The existence of redundancy in Convolutional Neural Networks (CNNs) enables us to remove some filters/channels with acceptable performance drops. However, the training objective of CNNs usually tends to minimize an accuracy-related loss function without any attention paid to the redundancy, making the redundancy distribute randomly on all the filters, such that removing any of them may trigger information loss and accuracy drop, necessitating a following finetuning step for recovery. In this paper, we propose to manipulate the redundancy during training to facilitate network pruning. To this end, we propose a novel Centripetal SGD (C-SGD) to make some filters identical, resulting in \textit{ideal redundancy patterns}, as such filters become \textit{purely redundant} due to their duplicates; hence removing them does not harm the network. As shown on CIFAR and ImageNet, C-SGD delivers better performance because the redundancy is better organized, compared to the existing methods. The efficiency also characterizes C-SGD because it is as fast as regular SGD, requires no finetuning, and can be conducted simultaneously on all the layers even in very deep CNNs. Besides, C-SGD can improve the accuracy of CNNs by first training a model with the same architecture but wider layers then squeezing it into the original width.
\end{abstract}

}

\maketitle

\IEEEdisplaynontitleabstractindextext

%
\IEEEpeerreviewmaketitle

\ifCLASSOPTIONcompsoc
\IEEEraisesectionheading{\section{Introduction}\label{sec:introduction}}
\else
\section{Introduction}
\label{sec:introduction}
\fi

%
%
%
%
\IEEEPARstart{C}{onvolutional} Neural Networks (CNNs) have become the de facto standard for computer vision and very deep architectures are much sought-after by visual tasks, such as image recognition, due to their approaching human-level performance. However, as CNNs grow wider and deeper, their memory footprint, power consumption and required floating-point operations (FLOPs) have increased dramatically, thus limiting their adoption within platforms without rich computational resource, like embedded systems. In this context, CNN compression and acceleration methods have been prevalent during the past few years, and the main research pathways include tensor low rank expansion~\cite{jaderberg2014speeding}, connection pruning~\cite{han2015learning}, filter pruning~\cite{li2016pruning}, quantization~\cite{han2015deep}, knowledge distillation~\cite{hinton2015distilling}, \etc. 

This paper focuses on filter pruning, a.k.a. channel pruning~\cite{he2017channel} or network slimming~\cite{liu2017learning}, because of its three unique features: \textbf{1)}~generic - it can handle various CNNs with no assumptions on the application field, the network architecture or the deployment platform; \textbf{2)} effective - it can significantly reduce the required FLOPs of the network, which serve as the main criterion of computational burdens; \textbf{3)} complementary to other techniques - it simply produces a thinner network with no customized structure or extra operation, which is orthogonal to the other model compression and acceleration methods.

In the past few years, tremendous efforts have been devoted to filter pruning techniques from both academia and industry. Due to the widely observed redundancy in CNNs~\cite{denil2013predicting,collins2014memory,cheng2015exploration,han2015deep,zhou2016less,yu2018nisp}, numerous works have shown that, if a CNN is pruned without a big decline in performance, a follow-up finetuning procedure may restore the performance to a certain degree. Some prior works~\cite{polyak2015channel,hu2016network,li2016pruning,molchanov2016pruning,abbasi2017structural,anwar2017structured} estimate the importance of filters by a variety of metrics, directly remove some filters and re-construct the network with the remaining ones. However, though the pruned filters are less important in some sense, they are not \textit{purely redundant}, hence the performance will be degraded. Moreover, some recent powerful networks adopt complicated structures, like shortcut~\cite{he2016deep} and dense connection~\cite{huang2017densely}, where some layers must be pruned in the same pattern as others, raising an open problem of \textit{constrained filter pruning}. This further challenges such pruning techniques, as the important filters at different layers usually reside in different positions, such that some important filters have to be pruned due to the constraints. In order to reduce the destructive impact of pruning, another family of methods~\cite{liu2015sparse,alvarez2016learning,wen2016learning,ding2018auto,wang2018structured,lin2019towards} seeks to zero out some filters in advance, where group-Lasso Regularization~\cite{roth2008group} is frequently used. The rationale behind is simple: the model undergoes less damage during pruning if the magnitudes of the pruned parameters have been reduced in advance. This is because pruning filters is mathematically equivalent to setting all of their parameters to zero. However, such regularizations cannot literally zero out the filters but merely reduce the magnitudes of their parameters to some extent (Sect.~\ref{sec-vs-zero-out}), hence the pruning operation still damages the model and a finetuning process remains necessary~\cite{liu2015sparse,alvarez2016learning,wen2016learning}. Essentially, zeroing filters out can be regarded as producing a \textit{redundancy pattern}, which we refer to as \textit{small-norm redundancy} for convenience. As some filters become more redundant (\ie, smaller in magnitude) than before but still not purely redundant, small-norm redundancy pattern is non-ideal.

In this paper, we also aim to produce some redundancy patterns in CNNs for filter pruning. However, unlike the non-ideal small-norm redundancy pattern, we seek to produce ideal patterns, where some filters are \textit{purely redundant}, such that the removal of them is not harmful to the model. To this end, we intend to merge multiple filters into one, thus generating a redundancy pattern where some filters are \textit{identical}. In the meantime, supervised by the model's original objective function, the performance is maintained. Compared to the importance-based filter pruning methods, doing so requires no heuristic knowledge about the importance of a filter. In contrast to the small-norm methods, the redundancy pattern is ideal, which enables absolutely lossless pruning and eliminates the need for a time-consuming finetuning process. 

\begin{figure*}
	\begin{center}
		\centerline{\includegraphics[width=\linewidth]{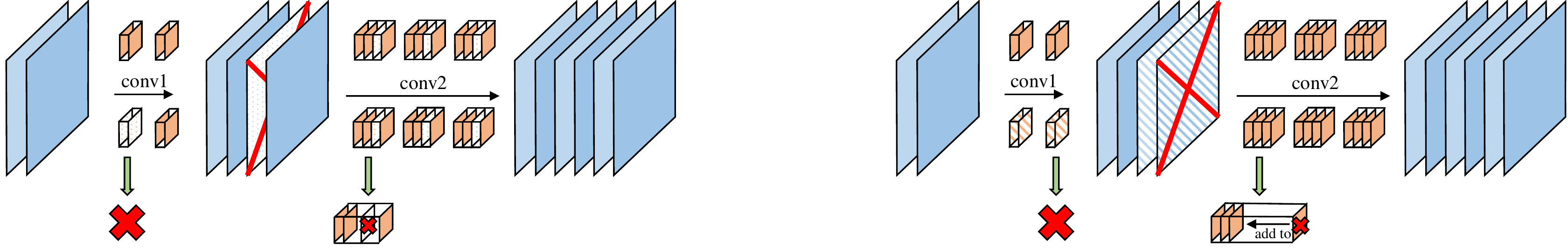}}
		\caption{Zeroing-out \vs centripetal constraint. This figure shows a CNN with 4 and 6 filters at the 1st and 2nd convolutional layer, respectively, which takes a 2-channel input. Left: the 3rd filter at conv1 is zeroed out (\ie, all the entries in the parameter tensor of the 3rd filter are close to zero, more precisely, $\bm{K}^{(1)}_{:,:,:,3}\approx\bm{0}$ using the formulation described in Sect. \ref{sec-formulation}), thus the 3rd feature map channel is close to zero ($\bm{M}^{(1)}_{:,:,3}\approx\bm{0}$), implying that the 3rd input channels of the 6 filters at conv2 are useless. During pruning, the 3rd filters at conv1 and the 3rd input channels of the 6 filters at conv2 are removed. Right: the 3rd and 4th filters at conv1 are forced to grow close by the centripetal constraint until the 3rd and 4th feature map channels become identical. But the 3rd and 4th input channels of the 6 filters at conv2 can still grow without constraints, making the encoded information still in full use. When pruned, the 4th filter at conv1 is removed, and the 4th input channel of every filter at conv2 is added to the 3rd channel.}
		\label{motivation-sketch}
	\end{center}
	\vskip -0.2in
\end{figure*}

\begin{figure*}
	\begin{center}
		\centerline{\includegraphics[width=\linewidth]{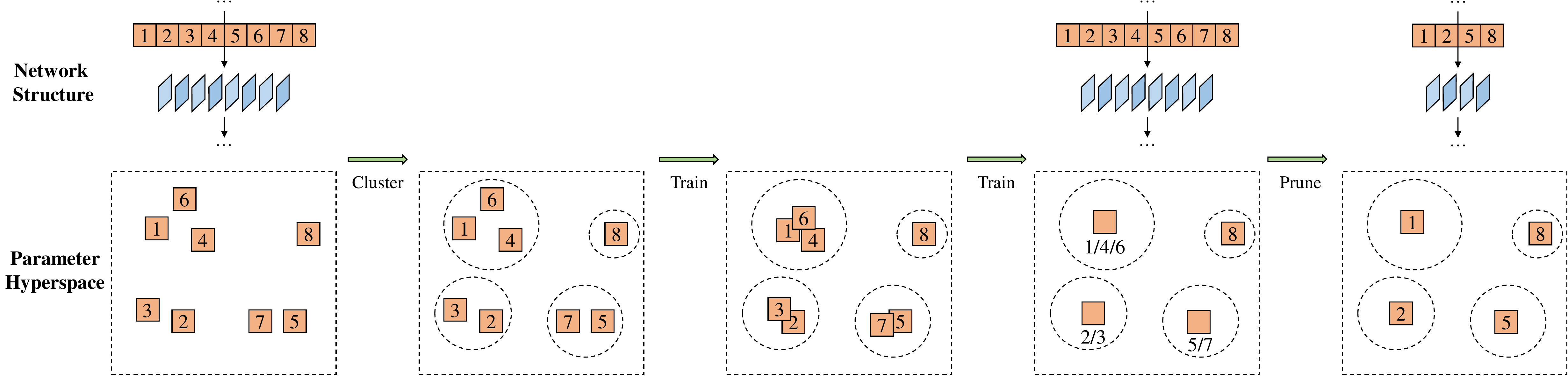}}
		\caption{The C-SGD pruning pipeline. When we seek to slim a 8-filter convolutional layer down to 4 filters, we divide the filters into 4 clusters according to their location in the parameter hyperspace. For instance, for a $3\times3$ convolutional layer which takes a $64$-channel feature map as input, every filter kernel has $3\times3\times64=576$ parameters, thus the dimensionality of hyperspace is 576. During C-SGD training, the filters in each cluster become closer and eventually identical. When the training is completed, we remove all but one filter for each cluster, and adjust the following layer as Fig. \ref{motivation-sketch}.}
		\label{intuition-sketch}
	\end{center}
	\vskip -0.2in
\end{figure*}

The intuition motivating the proposed method is an observation of the information flow in CNNs, as shown in Fig. \ref{motivation-sketch}. It reveals \textbf{1)} if two or more filters are trained to become identical, due to the \textit{linearity} of convolution, we can simply discard all but leave one filter, and add up the parameters along the corresponding input channels of the next layer. Doing so will lead to ZERO damage on performance; \textbf{2)} by encouraging multiple filters to grow closer in the parameter hyperspace, which we refer to as the \textit{centripetal constraint}, though they start to produce increasingly similar information, the information conveyed from the corresponding input channels of the next layer is still in full use. Therefore, the representational capacity of our model is probably weaker than that of the original expensive model, but stronger than a counterpart with the filters being zeroed out (Sect. \ref{sec-vs-zero-out}), as the input channels corresponding to the zeroed-out filters no longer contribute to the information flow~\cite{wen2016learning}. On the other hand, we will show that compared with a model without any manipulated redundancy, training with identical filters delivers higher accuracy (Sect. \ref{sec-redundant-train}). Our codes are released at \url{https://github.com/DingXiaoH/Centripetal-SGD} to encourage further studies. Our contributions are summarized as follows:
\begin{itemize}[noitemsep,nolistsep,,topsep=0pt,parsep=0pt,partopsep=0pt]
	\item We propose to produce ideal redundancy patterns in CNNs by training some filters to become identical (Fig. \ref{intuition-sketch}) via \textit{Centripetal SGD} (C-SGD), an efficient SGD optimization method which can solve \textit{constrained filter pruning}. Here ``centripetal'' means ``several objects moving towards a center'', which describes the behavior of the filters in C-SGD.
	\item We present an efficient implementation of C-SGD with matrix multiplications, which introduces no observable computational burdens, compared to normal SGD.
	\item As our theoretical contribution, we show that training a model with identical filters using C-SGD \textit{from scratch} delivers higher accuracy than a counterpart without such redundancy. This serves as evidence for supporting our motivation (Fig.~\ref{motivation-sketch}) as well as the assumption that redundancy helps the convergence of neural networks~\cite{denton2014exploiting,hinton2015distilling}.
	\item We propose a novel approach, Scaling and Squeezing, to improve the accuracy of CNNs, which first trains a model with the same architecture but wider layers and then squeezes it into the original width via C-SGD. Compared to the prior methods which fail to utilize the weights inherited from a wider model by pruning and finetuning, Scaling and Squeezing can improve the performance by a clear margin.
	\item Our experiments on CIFAR-10 and ImageNet have beaten many recent competitors by a clear margin in pruning several benchmark models. Our results on COCO detection and VOC segmentation demonstrate the generalization performance of C-SGD on the downstream tasks.
\end{itemize}


\section{Related Work}
\subsection{Filter Pruning}
Numerous inspiring researches
\cite{lecun1990optimal,hassibi1993second,castellano1997iterative,stepniewski1997pruning,han2015learning,guo2016dynamic,zhang2018systematic} have shown that it is feasible to remove a large portion of connections (\ie, weights) from a neural network without a significant performance drop. However, as such methods do not make the parameter tensors smaller but just sparser, little or no acceleration can be observed without the support from specialized software and hardware platforms. Then it is natural for researchers to go further on CNNs: by removing filters instead of sporadic connections, we transform the wide convolutional layers into narrower ones, hence the FLOPs, memory footprint and power consumption are significantly reduced. One kind of methods estimates the importance of filters by some means, then selects and prunes the unimportant filters carefully to minimize the performance loss. Some prior works measure a filter's importance by the classification accuracy reduction (CAR)~\cite{abbasi2017structural}, the channel contribution variance~\cite{polyak2015channel}, the Taylor-expansion criterion~\cite{molchanov2016pruning}, the magnitude of convolution kernels~\cite{li2016pruning} and the average percentage of zero activations (APoZ)~\cite{hu2016network}, respectively; Luo \etal~\cite{luo2017thinet} select filters based on the information derived from the next layer; Yu \etal~\cite{yu2018nisp} take into consideration the effect of error propagation; He \etal~\cite{he2017channel} select filters by solving the Lasso regression; He \etal~\cite{he2018adc} pick up filters with the aid of reinforcement learning. Another category seeks to train the network under certain constraints to zero out some filters~\cite{liu2015sparse,alvarez2016learning,wen2016learning,ding2018auto,wang2018structured,lin2019towards}, where the representative is group-Lasso Regularization.

Some major drawbacks of the prior works are as follows. \textbf{1)}~For the importance-based methods, the filter importance metrics are essentially heuristic, as it is not clear why the proposed metrics reflect the inherent importance of filters. Also, it is hard to judge if a heuristic metric is theoretically better than another. \textbf{2)} Since removing whole filters can degrade the performance a lot, the models are usually pruned in a layer-by-layer~\cite{polyak2015channel,alvarez2016learning,hu2016network} or filter-by-filter~\cite{molchanov2016pruning,abbasi2017structural} manner. Running on today's very deep CNNs, such pruning processes may not only become time-consuming but also suffer from the notorious problem of error propagation and amplification through multiple layers when estimating the filter importance~\cite{yu2018nisp}. \textbf{3)} Many of these works require one or more finetuning processes after pruning to restore the accuracy~\cite{polyak2015channel,alvarez2016learning,hu2016network,molchanov2016pruning,wen2016learning}. However, Liu \etal~\cite{liu2018rethinking} have empirically found out that finetuning a pruned model may not always guarantee higher accuracy, compared to training from scratch, as the pruned model might be trapped into a bad local minimum. \textbf{4)} The regularization-based methods may bring significant extra computational burdens. For example, in our experiments (Sect.~\ref{sec-vs-zero-out}), group-Lasso Regularization~\cite{roth2008group} on 3/8 of the filters slows down the training by about $2\times$, as it requires costly square root operations. \textbf{5)} Many of the methods cannot handle the constrained filter pruning problem on ResNets (Fig. \ref{fig-res-break}), so the researchers choose to sidestep this problem by only pruning the internal layers in residual blocks~\cite{li2016pruning,he2017channel,liu2017learning}. Li \etal~\cite{li2016pruning} and Ding \etal~\cite{ding2018auto} tried pruning the troublesome layers according to the importance scores of others in order to meet the constraints, but predictably resulted in significantly lower accuracy.

In contrast, our method features \textbf{1)} no heuristic knowledge about the filter importance, \textbf{2)} the capability of pruning every target layer simultaneously, \textbf{3)} no need for finetuning, \textbf{4)} negligible extra computations, and \textbf{5)} global slimming on all the layers in complicated CNN architectures.

This paper represents a very substantial extension of our previous conference paper~\cite{CSGD}. The main technical novelties, compared with~\cite{CSGD}, are as follows. \textbf{1)} We propose a novel CNN training methodology, Scaling and Squeezing, in order to improve the performance of CNN based on C-SGD. \textbf{2)} We present more experimental results including the pruning results on VGG \cite{simonyan2014very}, the torchvision \cite{torch-model} version of ResNet-50, and the object detection and semantic segmentation results on COCO and VOC. \textbf{3)} We present more illustrations and discussions of the motivation and derivation of C-SGD together with its relation to the prior works. \textbf{4)} We present thorough comparisons and discussions of different clustering methods, including k-means, even and imbalanced clustering, on several benchmark models. \textbf{5)} We perform controlled experiments to justify the significance of solving the constrained filter pruning problem. \textbf{6)} We present more detailed discussions of the efficiency of C-SGD and its applications.

\subsection{Other Methods}
Apart from filter pruning, some works compress and accelerate CNNs in other ways.  Considerable works~\cite{sainath2013low,xue2013restructuring,denton2014exploiting,jaderberg2014speeding,kim2015compression,sindhwani2015structured,zhang2016accelerating,alvarez2017compression} decompose or approximate parameter tensors; quantization techniques~\cite{gupta2015deep,han2015deep,rastegari2016xnor,wu2016quantized,courbariaux2016binarized} approximate a model using fewer bits per parameter; knowledge distillation methods~\cite{ba2014deep,romero2014fitnets,hinton2015distilling} transfer knowledge from a big network to a smaller one; some researchers seek to speed up convolution with the help of perforation~\cite{figurnov2016perforatedcnns}, FFT~\cite{mathieu2013fast,vasilache2014fast} or DCT~\cite{wang2016cnnpack}; Wang \etal~\cite{wang2017beyond} compact feature maps by extracting information via a Circulant matrix. And some compression techniques can be combined to achieve the desired trade-off. E.g., a well-known preceding work named Deep Compression~\cite{han2015deep} noticed and significantly reduced the redundancy of deep CNN by connection pruning, quantization and Huffman encoding, but no filter pruning.

Of note is that C-SGD-based filter pruning is \textit{orthogonal} to these techniques described above, as it simply shrinks a wide CNN into a narrower one with no special structures or extra operations.

\section{Filter Pruning via Centripetal SGD} 
\subsection{Formulation}\label{sec-formulation}
In modern CNNs, batch normalization~\cite{ioffe2015batch} and linear scaling transformation commonly follow convolutional layers. For simplicity and generality, we regard a convolutional layer together with its subsequent batch normalization and scaling layer, if any, as a whole. Let $i$ be the layer index, $\bm{M}^{(i)}\in\mathbb{R}^{h_{i}\times w_{i}\times c_{i}}$ be the output feature map of layer $i$ with a spatial resolution of $h_i\times w_i$ and $c_{i}$ channels, and $\bm{M}^{(i)}_{:,:,j}$ be the $j$-th channel. The convolutional layer $i$ with kernel size $u_i\times v_i$ has one 4th-order tensor and four vectors as parameters \textit{at most}, namely, $\bm{\mu}^{(i)}, \bm{\sigma}^{(i)}, \bm{\gamma}^{(i)}, \bm{\beta}^{(i)}\in\mathbb{R}^{c_{i}}$ and $\bm{K}^{(i)}\in \mathbb{R}^{u_{i}\times v_{i}\times c_{i-1}\times c_{i}}$, where $\bm{K}^{(i)}$ is the convolution kernel, $\bm{\mu}^{(i)}$ and $\bm{\sigma}^{(i)}$ are the mean and standard deviation of batch normalization, $\bm{\gamma}^{(i)}$ and $\bm{\beta}^{(i)}$ are the scaling factor and bias term of the linear transformation, respectively. Then we use $\bm{P}^{(i)}=(\bm{K}^{(i)},\bm{\mu}^{(i)},\bm{\sigma}^{(i)},\bm{\gamma}^{(i)},\bm{\beta}^{(i)})$ to denote the parameters of layer $i$. In this paper, a filter $j$ at layer $i$ refers to the five-tuple comprising all the parameter slices related to the output channel $j$ of layer $i$, formally,
\begin{equation}
\bm{F}^{(j)}=(\bm{K}^{(i)}_{:,:,:,j},\mu^{(i)}_j,\sigma^{(i)}_j,\gamma^{(i)}_j,\beta^{(i)}_j) \,,
\end{equation}
where $\bm{K}^{(i)}_{:,:,:,j}$ is the $j$-th slice along the axis which differentiates the $c$ filters, \ie, the 4th axis in our formulation.

This layer takes $\bm{M}^{(i-1)}\in\mathbb{R}^{h_{i-1}\times w_{i-1}\times c_{i-1}}$ as input and outputs $\bm{M}^{(i)}$. Let $\ast$ be the 2-D convolution operator, the $j$-th output channel is
\begin{equation}\label{def-convolution}
\bm{M}^{(i)}_{:,:,j}=\frac{\sum_{k=1}^{c_{i-1}}\bm{M}^{(i-1)}_{:,:,k}\ast\bm{K}^{(i)}_{:,:,k,j}-\mu^{(i)}_j}{\sigma^{(i)}_j}\gamma^{(i)}_j+\beta^{(i)}_j \,.
\end{equation}

Pruning filters at a certain layer normally conducts the following three steps: \textbf{1)} deciding which filters to prune, \textbf{2)} deleting the corresponding parameters in the kernel, \eg, along the 4th axis in our formulation, and \textbf{3)} handling the vector parameters $\bm{\mu},\bm{\sigma},\bm{\gamma}$ and $\bm{\beta}$ accordingly. In practice, we equivalently reconstruct a new network with a narrower layer, and use the parameters of the remaining filters in the original model for initialization.

For example, the importance-based filter pruning methods~\cite{polyak2015channel,hu2016network,li2016pruning,molchanov2016pruning,abbasi2017structural,yu2018nisp} define the importance of filters by some means to guide the selection of important filters. Let $\mathcal{I}_i$ be the filter index set of layer $i$ (\eg, $\mathcal{I}_2=\{1,2,3,4\}$ if the 2nd layer has 4 filters), $T$ be the filter importance evaluation function and $\theta_i$ be the threshold, the \textit{remaining set}, \ie, the index set of the filters which survive the pruning, is 
\begin{equation}
\mathcal{R}_i=\{j\in \mathcal{I}_i \ |\ T(\bm{F}^{(j)})>\theta_i\} \,.
\end{equation}

We construct the parameters for the slimmed layer by assembling the parameters sliced from the original tensor and vectors into the new parameters $\hat{\bm{P}}^{(i)}$. That is,
\begin{equation}\label{eq4}
\hat{\bm{P}}^{(i)}=(\bm{K}^{(i)}_{:,:,:,\mathcal{R}_i},\bm{\mu}^{(i)}_{\mathcal{R}_i},\bm{\sigma}^{(i)}_{\mathcal{R}_i},\bm{\gamma}^{(i)}_{\mathcal{R}_i},\bm{\beta}^{(i)}_{\mathcal{R}_i}) \,.
\end{equation}

The input channels of the next layer corresponding to the pruned filters should also be discarded (Fig. \ref{motivation-sketch}),  
\begin{equation}\label{eq6}
\hat{\bm{P}}^{(i+1)}=(\bm{K}^{(i+1)}_{:,:,\mathcal{R}_i,:},\bm{\mu}^{(i+1)},\bm{\sigma}^{(i+1)},\bm{\gamma}^{(i+1)},\bm{\beta}^{(i+1)}) \,.
\end{equation}

Then we initialize the new network using $\hat{\bm{P}}^{(i)}$ and $\hat{\bm{P}}^{(i+1)}$.

\subsection{Centripetal SGD}
In this subsection, we present the update rule of C-SGD together with some discussions of its properties. An intuitive illustration will be provided in the following subsection.

For each layer, we first divide the filters into clusters, where the number of clusters equals the desired number of filters, as we preserve only one filter for each cluster. We use $\mathcal{C}_i$ and $\mathcal{H}$ to denote the set of all filter clusters of layer $i$ and a specific cluster in the form of a filter index set, respectively. We generate the clusters by k-means~\cite{hartigan1979algorithm} or \textit{arbitrarily}, between which our experiments demonstrate only minor difference (Sect. \ref{sec-exp-clustering-methods}). In the following sections, we use k-means clustering unless otherwise noted.
\begin{itemize}[noitemsep,nolistsep,,topsep=0pt,parsep=0pt,partopsep=0pt]
	\item \textbf{K-means clustering}. We aim to generate clusters with low intra-cluster distance in the parameter hyperspace, such that collapsing them into a single point less impacts the model, which is natural. To this end, we simply flatten the filter's kernel and use it as the feature vector for k-means clustering.
	\item \textbf{Even clustering}. We can generate clusters with no consideration of the filters' inherent properties. Let $c_i$ and $r_i$ be the number of original filters and desired remaining filters (\ie, number of clusters) at layer $i$, respectively, each cluster will have $\lceil c_i / r_i\rceil$ filters at most. For example, if the 2nd layer has 6 filters and we wish to slim it to 4 filters, we will have $\mathcal{C}_2=\{\mathcal{H}_1,\mathcal{H}_2,\mathcal{H}_3,\mathcal{H}_4\}$, $\mathcal{H}_1=\{1,2\},\mathcal{H}_2=\{3,4\},\mathcal{H}_3=\{5\},\mathcal{H}_4=\{6\}$.
	\item \textbf{Imbalanced clustering.} An extreme solution is to put $c_i - r_i + 1$ filters into one single cluster, such that each of the other clusters has only one filter. In the above example, we will have $\mathcal{H}_1=\{1,2,3\},\mathcal{H}_2=\{4\},\mathcal{H}_3=\{5\},\mathcal{H}_4=\{6\}$.
\end{itemize}

We use $H(j)$ to denote the cluster containing filter $j$, \eg, $H(3)=\mathcal{H}_1$ and $H(6)=\mathcal{H}_4$ in the above example of imbalanced clustering. Let $\bm{F}^{(j)}$ be the kernel or a vector parameter of filter $j$, at each training iteration, the update rule of C-SGD is
\begin{equation}\label{update-rule}
\begin{aligned}
\bm{F}^{(j)} \gets &\bm{F}^{(j)}+\tau \Delta\bm{F}^{(j)} \,, \\
\Delta\bm{F}^{(j)}=&-\frac{\sum_{k \in H(j)} \frac{\partial L}{\partial \bm{F}^{(k)}}}{|H(j)|} - \eta \bm{F}^{(j)} \\ &+ \epsilon (\frac{\sum_{k \in H(j)}\bm{F}^{(k)}}{|H(j)|}-\bm{F}^{(j)}) \,,
\end{aligned}
\end{equation}
where $L$ is the original objective function, $\tau$ is the learning rate, $\eta$ is the original weight decay factor, and $\epsilon$ is the only hyper-parameter we introduced, which is called the \textit{centripetal strength}.

The intuition behind Eq. \ref{update-rule} is quite simple: for the filters in the same cluster, the increments derived by the objective function are averaged (the first term), the normal weight decay is applied as well (the second term), and the difference in the initial values is gradually eliminated (the last term), so the filters will move towards their center in the hyperspace. 

Let $\mathcal{L}$ be the layer index set, \eg, $\mathcal{L}=\{1,2,3\}$ if the model has 3 convolutional layers, we use the \textit{sum of squared kernel deviation} $\chi$ to measure the intra-cluster similarity, \ie, how close filters are in each cluster,
\begin{equation}\label{eq-def-chi}
\chi=\sum_{i\in\mathcal{L}}\sum_{j\in \mathcal{I}_i}\Vert\bm{K}^{(i)}_{:,:,:,j} - \frac{\sum_{k\in H(j)}\bm{K}^{(i)}_{:,:,:,k}}{|H(j)|}\Vert_2^2 \,.
\end{equation}
It is easy to derive from Eq. \ref{update-rule} that $\chi$ is lowered \textit{monotonically} and \textit{exponentially} with a constant learning rate $\tau$, if the floating-point operation errors are ignored.

In practice, we fix $\eta$ and reduce $\tau$ with time just as we do in regular SGD training, and set $\epsilon$ \textit{casually}. Intuitively, C-SGD training with a large $\epsilon$ prefers rapid change to stable transition. If $\epsilon$ is too large, \eg, 10, the filters are merged immediately such that the whole process becomes equivalent to training a destroyed model from scratch. If $\epsilon$ is extremely small, like $1\times 10^{-10}$, the difference between C-SGD training and normal SGD is almost invisible during a long time. However, since the difference among filters in each cluster is reduced \textit{monotonically} and \textit{exponentially}, even an extremely small $\epsilon$ can make the filters close enough for absolutely lossless pruning, sooner or later. In this sense, we claim that such a redundancy pattern is ideal. As will be shown in Sect. \ref{sec-insensitive}, the performance of C-SGD is robust to the value of $\epsilon$.

\subsection{An Intuitive Illustration of C-SGD}
\begin{figure}[t]\label{fig-contour}
	\centering
	\subfloat[Normal weight decay.]
	{
		\includegraphics[width=0.47\linewidth,height=1.50in]{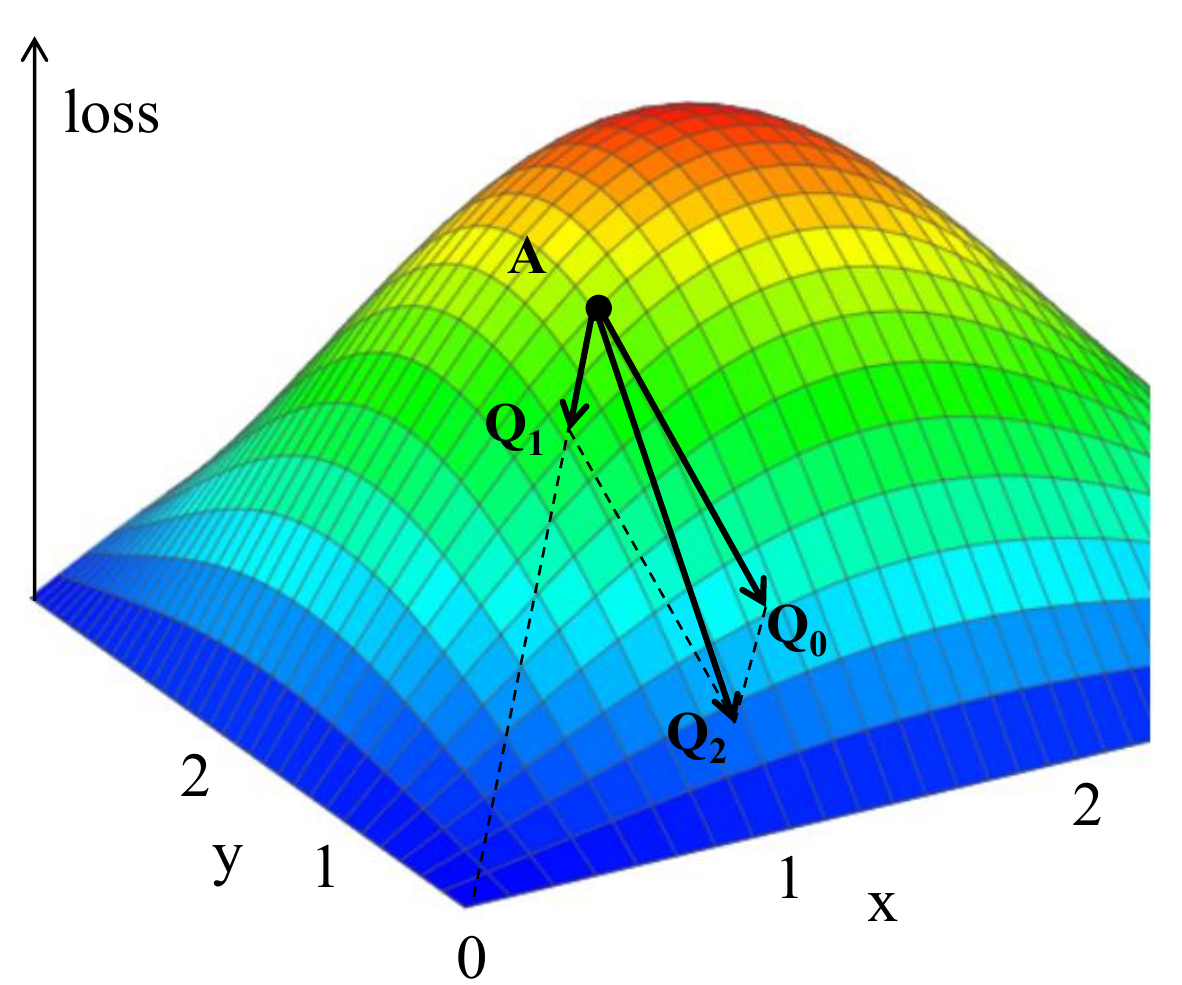}
		\label{fig-contour-weightdecay}
	}
	\subfloat[Centripetal constraint.]
	{
		\includegraphics[width=0.47\linewidth,height=1.50in]{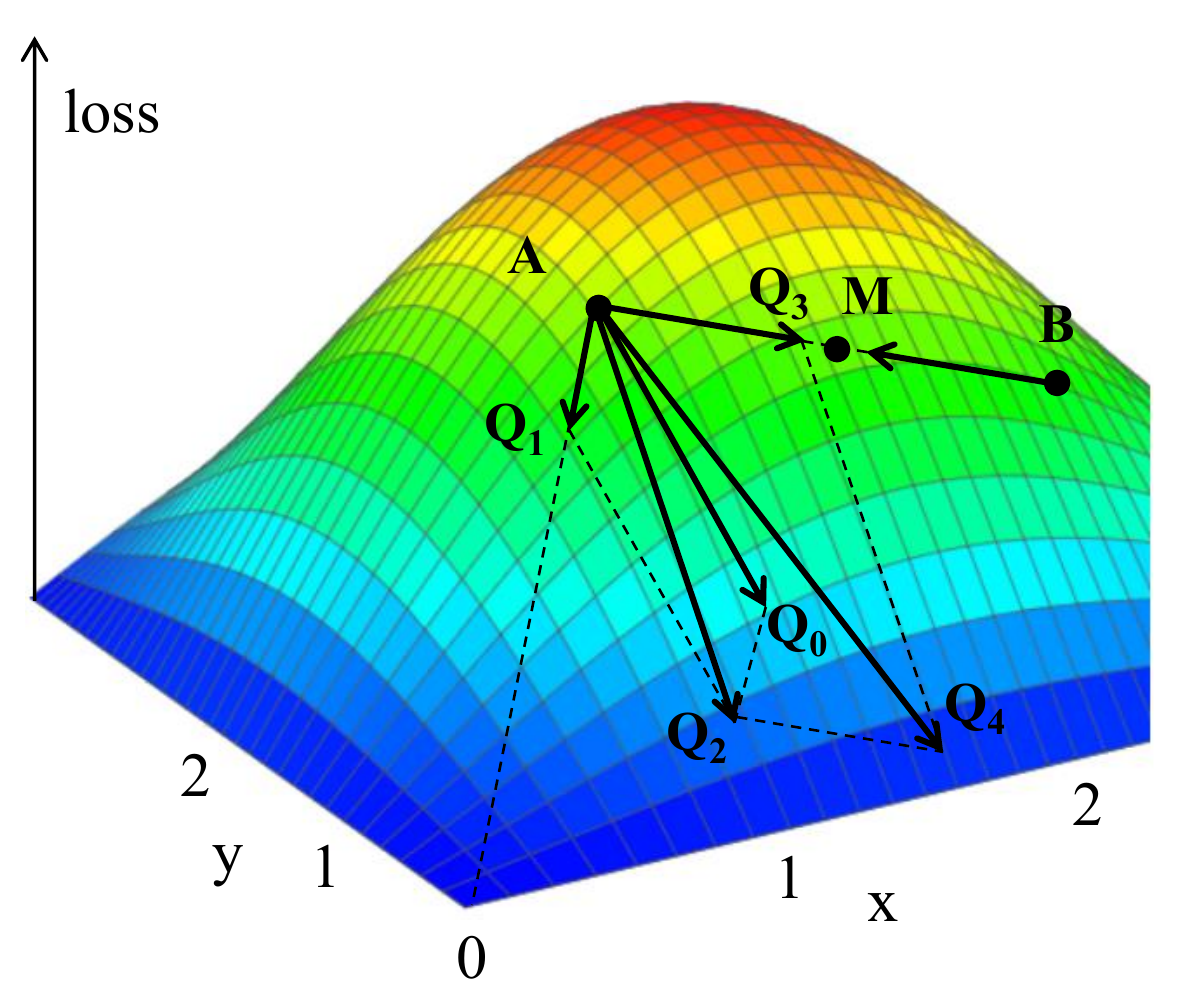}
		\label{fit-contour-csgd}
	}
	\caption{Gradient descent direction on the loss surface of regular weight decay and centripetal constraint without merging the original gradients.}
\end{figure}
A simple analogy to weight decay (\ie, $\ell$-2 regularization) may help understand Centripetal SGD. Fig. \ref{fig-contour-weightdecay} shows a 3-D loss surface, where a certain point $A$ corresponds to a 2-D parameter $\bm{a}=(a_1,a_2)$. Suppose the steepest descent direction is $\overrightarrow{AQ_0}$, we have $\overrightarrow{AQ_0}=-\frac{\partial L}{\partial\bm{a}}$, where $L$ is the objective function. Weight decay is commonly applied to reduce overfitting~\cite{krogh1992simple}, that is, $\overrightarrow{AQ_1}=-\eta\bm{a}$, where $\eta$ is the model's weight decay factor, \eg, $1\times 10^{-4}$ for ResNets~\cite{he2016deep}. The actual gradient descent direction then becomes $\Delta\bm{a}=\overrightarrow{AQ_2}=\overrightarrow{AQ_0}+\overrightarrow{AQ_1}=-\frac{\partial L}{\partial\bm{a}}-\eta\bm{a}$.

If we wish to make $A$ and $B$ closer to each other than they used to be, a natural idea is to push both $A$ and $B$ towards their midpoint $M(\frac{\bm{a} + \bm{b}}{2})$, as shown in Fig. \ref{fit-contour-csgd}. The gradient descent direction of point $A$ becomes 
\begin{equation}\label{delta-a}
\Delta\bm{a}=\overrightarrow{AQ_2}+\overrightarrow{AQ_3}=-\frac{\partial L}{\partial\bm{a}}-\eta\bm{a}+\epsilon(\frac{\bm{a}+\bm{b}}{2}-\bm{a}) \,,
\end{equation}
where $\epsilon$ is a hyper-parameter controlling the intensity or speed of pushing $A$ and $B$ close. Similarly, 
\begin{equation}\label{delta-b}
\Delta\bm{b}=-\frac{\partial L}{\partial\bm{b}}-\eta\bm{b}+\epsilon(\frac{\bm{a}+\bm{b}}{2}-\bm{b}) \,.
\end{equation}

Further, we wish to force point $A$ and $B$ to become increasingly close and \textit{eventually the same}. Formally, let $t$ be the number of training iterations, we aim to achieve
\begin{equation}
\lim\limits_{t\to\infty}\Vert\bm{a}^{(t)}-\bm{b}^{(t)}\Vert=0 \,,
\end{equation}
which implies
\begin{equation}
\lim\limits_{t\to\infty}\Vert\bm{a}^{(t+1)}-\bm{b}^{(t+1)}\Vert=0 \,,
\end{equation}
or
\begin{equation}\label{lim-satisfy}
\lim\limits_{t\to\infty}\Vert(\bm{a}^{(t)}-\bm{b}^{(t)})+\tau(\Delta\bm{a}^{(t)}-\Delta\bm{b}^{(t)})\Vert=0 \,,
\end{equation}
since $\bm{a}^{(t+1)}=\bm{a}^{(t)}+\tau\Delta\bm{a}^{(t)}$ and $\bm{b}^{(t+1)}=\bm{b}^{(t)}+\tau\Delta\bm{b}^{(t)}$, where $\tau$ is the learning rate. We seek to achieve Eq. \ref{lim-satisfy} by both
\begin{equation}\label{lim-satisfy-grad}
\lim\limits_{t\to\infty}(\Delta\bm{a}^{(t)}-\Delta\bm{b}^{(t)})=\bm{0} \,,
\end{equation}
\begin{equation}
\lim\limits_{t\to\infty}(\bm{a}^{(t)}-\bm{b}^{(t)})=\bm{0} \,.
\end{equation}

\begin{figure}
	\begin{center}
		\centerline{\includegraphics[width=0.5\linewidth]{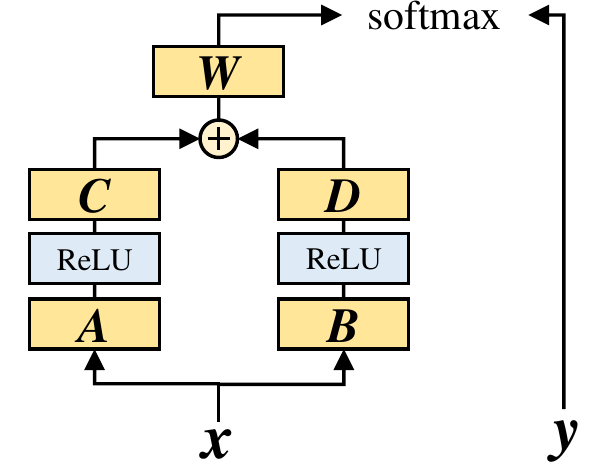}}
		\caption{An example explaining that equal parameters do not imply equal gradients. Let $L$ be the loss function of a simple network, $L(\bm{x},\bm{y})=\text{softmax}(\bm{W}(\bm{C}(\text{ReLU}(\bm{Ax})) + \bm{D}(\text{ReLU}(\bm{Bx}))), \bm{y})$, $\bm{A}=\bm{B}$ does not imply $\frac{\partial L}{\partial\bm{A}} = \frac{\partial L}{\partial\bm{B}}$, because $\bm{C}\neq\bm{D}$. In the case of two filters in CNNs, their outputs (feature map channels) are convolved by different kernels at the subsequent layer, so we cannot ensure the equality of their gradients using normal SGD.}
		\label{fig-grad-example}
	\end{center}
\vskip -0.2in
\end{figure}

Namely, as two points are growing closer, their gradients should become closer accordingly for the training to converge. But here we encounter a problem: given Eq. \ref{delta-a} and Eq. \ref{delta-b}, we have 
\begin{equation}
\Delta\bm{a}^{(t)}-\Delta\bm{b}^{(t)} = (\frac{\partial L}{\partial\bm{b}^{(t)}} -\frac{\partial L}{\partial\bm{a}^{(t)}}) + (\eta+\epsilon)(\bm{b}^{(t)}-\bm{a}^{(t)}) \,,
\end{equation}
but cannot ensure $\lim\limits_{t\to\infty}(\frac{\partial L}{\partial\bm{b}^{(t)}}-\frac{\partial L}{\partial\bm{a}^{(t)}})=\bm{0}$. Actually, as shown in Fig. \ref{fig-grad-example}, even $\bm{a}=\bm{b}$ does not imply $\frac{\partial L}{\partial\bm{a}}=\frac{\partial L}{\partial\bm{b}}$, so Eq. \ref{lim-satisfy-grad} cannot be met with Eq. \ref{delta-a} and Eq. \ref{delta-b}.

We solve this problem by merging the gradients derived from the original objective function. For simplicity and symmetry, we replace both $\frac{\partial L}{\partial\bm{a}}$ in Eq. \ref{delta-a} and $\frac{\partial L}{\partial\bm{b}}$ in Eq. \ref{delta-b} by $\frac{1}{2}(\frac{\partial L}{\partial\bm{a}}+\frac{\partial L}{\partial\bm{b}})$. This way ensures the supervision information encoded in the objective-function-related gradients to be preserved so as to maintain the model's performance, and Eq. \ref{lim-satisfy} to be met, which can be easily verified. Intuitively, we deviate $\bm{a}$ from the steepest descent direction according to the information of $\bm{b}$ and deviate $\bm{b}$ vice versa, just like the normal $\ell$-2 regularization deviates both $\bm{a}$ and $\bm{b}$ towards the origin of coordinates. In summary, we have
\begin{equation}
\Delta\bm{a} = -\frac{1}{2}(\frac{\partial L}{\partial\bm{a}}+\frac{\partial L}{\partial\bm{b}}) -\eta\bm{a}+\epsilon(\frac{\bm{a}+\bm{b}}{2}-\bm{a}) \,,
\end{equation}
\begin{equation}
\Delta\bm{b} = -\frac{1}{2}(\frac{\partial L}{\partial\bm{a}}+\frac{\partial L}{\partial\bm{b}}) -\eta\bm{b}+\epsilon(\frac{\bm{a}+\bm{b}}{2}-\bm{b}) \,,
\end{equation}
which is exactly the degraded version of Eq. \ref{update-rule}.

\subsection{Efficient Implementation of C-SGD}
The efficiency of modern CNN training and deployment platforms, \eg, Tensorflow~\cite{abadi2016tensorflow}, is dependent on large-scale tensor operations. We therefore seek to implement C-SGD by efficient matrix multiplications which introduce minimal computational burdens. Concretely, given a convolutional layer $i$, the kernel $\bm{K}\in\mathbb{R}^{u_i\times v_i\times c_{i-1} \times c_i}$ and the gradient $\frac{\partial L}{\partial \bm{K}}$, we reshape $\bm{K}$ to $\bm{W}\in\mathbb{R}^{u_i v_i c_{i-1} \times c_i}$ and $\frac{\partial L}{\partial \bm{K}}$ to $\frac{\partial L}{\partial \bm{W}}$ accordingly. We construct the averaging matrix $\bm{\Gamma}\in\mathbb{R}^{c_i \times c_i}$ and decaying matrix $\bm{\Lambda}\in\mathbb{R}^{c_i \times c_i}$ as Eq. \ref{def-avg-matrix} and Eq. \ref{def-decay-matrix} such that Eq. \ref{matrix-mul} is equivalent to Eq. \ref{update-rule}, which can be easily verified. Obviously, when the number of clusters equals that of the filters, Eq. \ref{matrix-mul} degrades into normal SGD with $\bm{\Gamma}=\text{diag}(1), \bm{\Lambda}=\text{diag}(\eta)$. The other trainable parameters (\ie, $\bm{\gamma}$ and $\bm{\beta}$) are reshaped into $\bm{W}\in\mathbb{R}^{1 \times c_i}$ and handled in the same way. In practice, we observe almost no difference in the speed between normal SGD and C-SGD using Tensorflow on Nvidia GeForce GTX 1080Ti GPUs with CUDA9.0 and cuDNN7.0.
\begin{equation}\label{matrix-mul}
\bm{W} \gets \bm{W}-\tau (\frac{\partial L}{\partial \bm{W}}\bm{\Gamma} + \bm{W}\bm{\Lambda}) \,.
\end{equation}
\begin{equation}\label{def-avg-matrix}
\bm{\Gamma}_{m,n}=
\begin{dcases}
\frac{1}{|H(m)|} & \text{if $H(m)=H(n)$} \,, \\
0 & \text{elsewise} \,.
\end{dcases}
\end{equation}
\begin{equation}\label{def-decay-matrix}
\bm{\Lambda}_{m,n}=
\begin{dcases}
\eta + \epsilon - \frac{\epsilon}{|H(m)|} & \text{if $m=n$} \,, \\
- \frac{\epsilon}{|H(m)|} & \text{if $m\neq n$, $H(m)=H(n)$} \,, \\
0 & \text{elsewise} \,.
\end{dcases}
\end{equation}

\subsection{Filter Trimming after C-SGD Training}
When the training is completed, since the ideal redundancy patterns have emerged, \ie, the filters in each cluster have become identical, picking up which one makes no difference. We simply pick up the first filter (\ie, the filter with the smallest index) in each cluster to form the remaining set for each layer,
\begin{equation}
\mathcal{R}_i=\{\text{min}(\mathcal{H}) \ |\ \forall \mathcal{H} \in \mathcal{C}_i\} \,.
\end{equation}

For the following layer, we add the to-be-deleted input channels to the corresponding remaining one (Fig. \ref{motivation-sketch}),
\[
\bm{K}^{(i+1)}_{:,:,k,:}\gets\sum \bm{K}^{(i+1)}_{:,:,H(k),:} \quad \forall k \in \mathcal{R}_i \,,
\]
then we delete the redundant filters as well as the input channels of the following layer as Eq. \ref{eq4}, \ref{eq6}. Due to the linearity of convolution (Eq. \ref{def-convolution}), no damage is caused, hence \textit{no finetuning} is needed.

\subsection{C-SGD for Constrained Filter Pruning}
\begin{figure}
	\begin{center}
		\centerline{\includegraphics[width=0.97\linewidth]{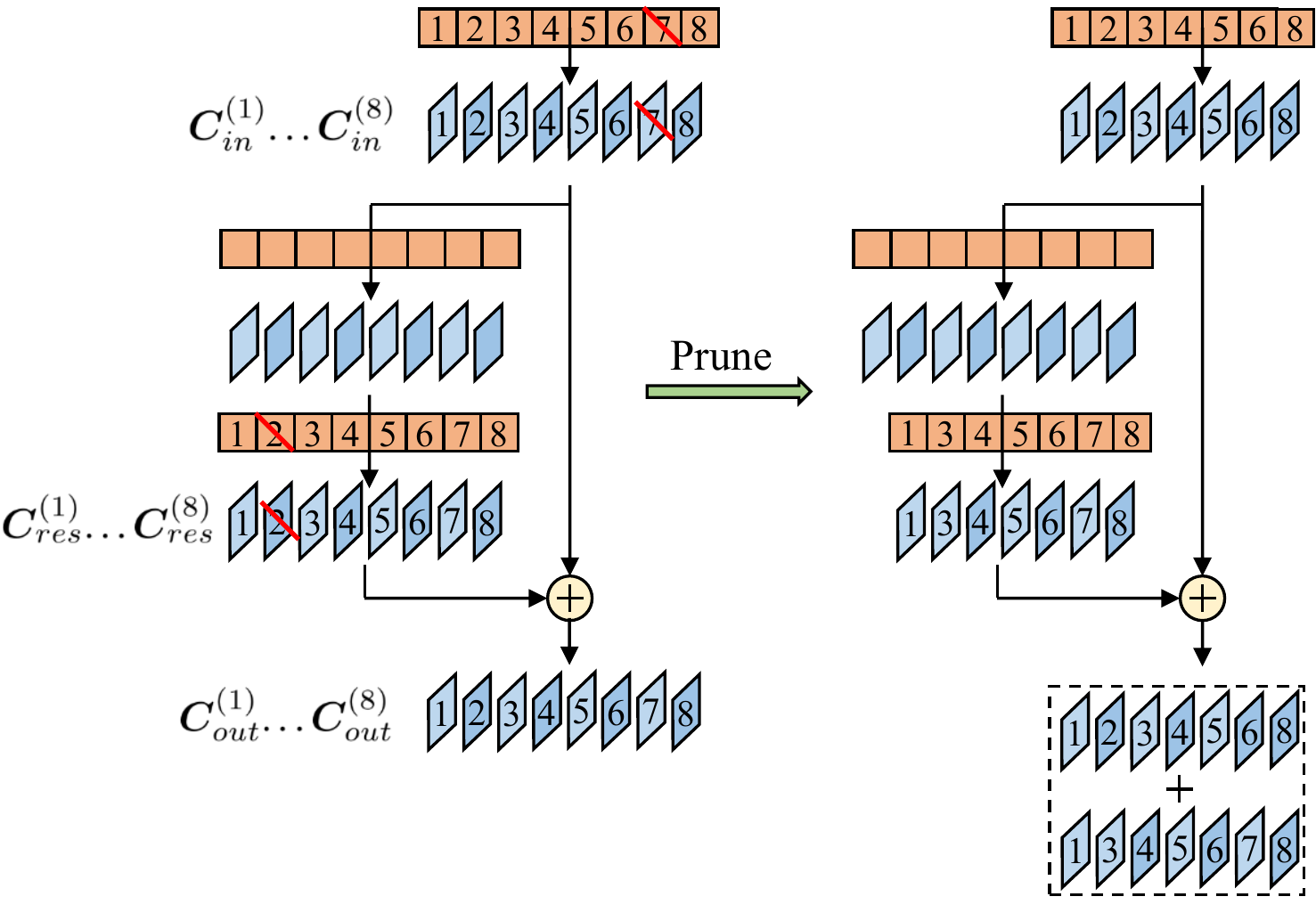}}
		\caption{We explain the problem of constrained filter pruning in a ResNet where an 8-channel featuremap (denoted by $\bm{C}_{in}^{(i)} \in \mathbb{R}^{h\times w}, 1\leq i \leq 8$) is fed into a residual block, and the output of the last layer of residual block $\bm{C}_{res}^{(i)}$ is added onto $\bm{C}_{in}^{(i)}$. Here we have $\bm{C}_{out}^{(i)} = \bm{C}_{in}^{(i)} + \bm{C}_{res}^{(i)}, \forall 1\leq i \leq 8$. By filter pruning, we expect to remove one or several channels without breaking the correspondence between the remaining channels which are added up. Therefore, the remaining sets of the first and last layer must be the same. Otherwise, if we prune the 7th filter at the first layer and the 2nd filter at the last layer of residual block for example, we will end up with $\bm{C}_{out}^{(2)} = \bm{C}_{in}^{(2)} + \bm{C}_{res}^{(3)}, \bm{C}_{out}^{(6)} = \bm{C}_{in}^{(6)} + \bm{C}_{res}^{(7)}$ \dots Hence the correspondence breaks, and the model is destroyed.}
		\label{fig-res-break}
	\end{center}
\end{figure}
\begin{figure}[t]
	\centering
	\subfloat[Original.]
	{
		\includegraphics[page=1,width=0.22\linewidth]{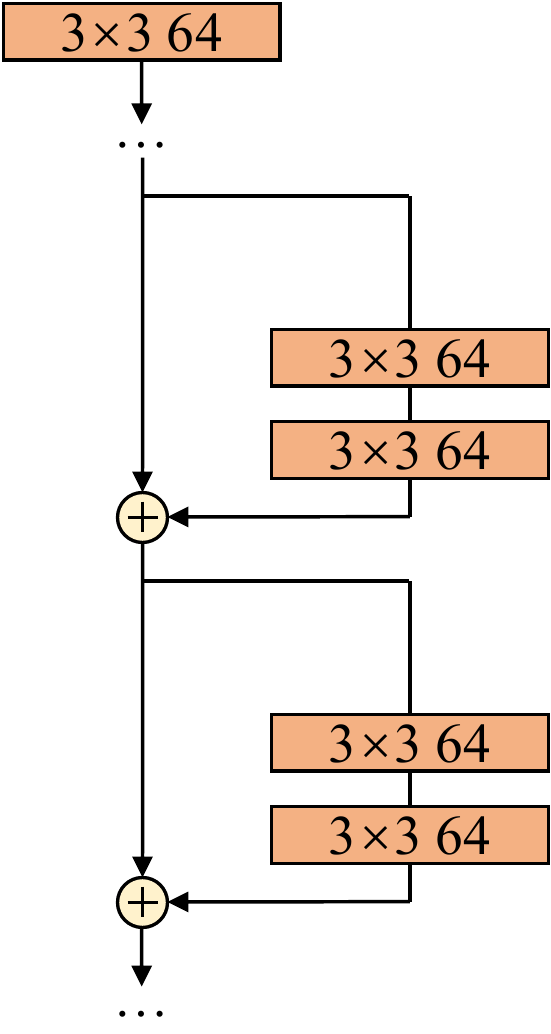} 
	}
	\subfloat[Clipped.]
	{
		\includegraphics[page=2,width=0.22\linewidth]{literal_slim.pdf} 
	}
	\subfloat[Sampled.]
	{
		\includegraphics[page=3,width=0.22\linewidth]{literal_slim.pdf} 
	}
	\subfloat[Slimmed.]
	{
		\includegraphics[page=4,width=0.22\linewidth]{literal_slim.pdf} 
	}
	\caption{Compared to prior works which only clip the internal layers~\cite{li2016pruning} or insert sampler layers~\cite{he2017channel,liu2017learning} on ResNets, C-SGD literally ``slims'' the network.}
	\label{fig-slim-clip}
\end{figure}
\begin{figure*}
	\begin{center}
		\setlength{\abovecaptionskip}{5pt} 
		\centerline{\includegraphics[width=\linewidth]{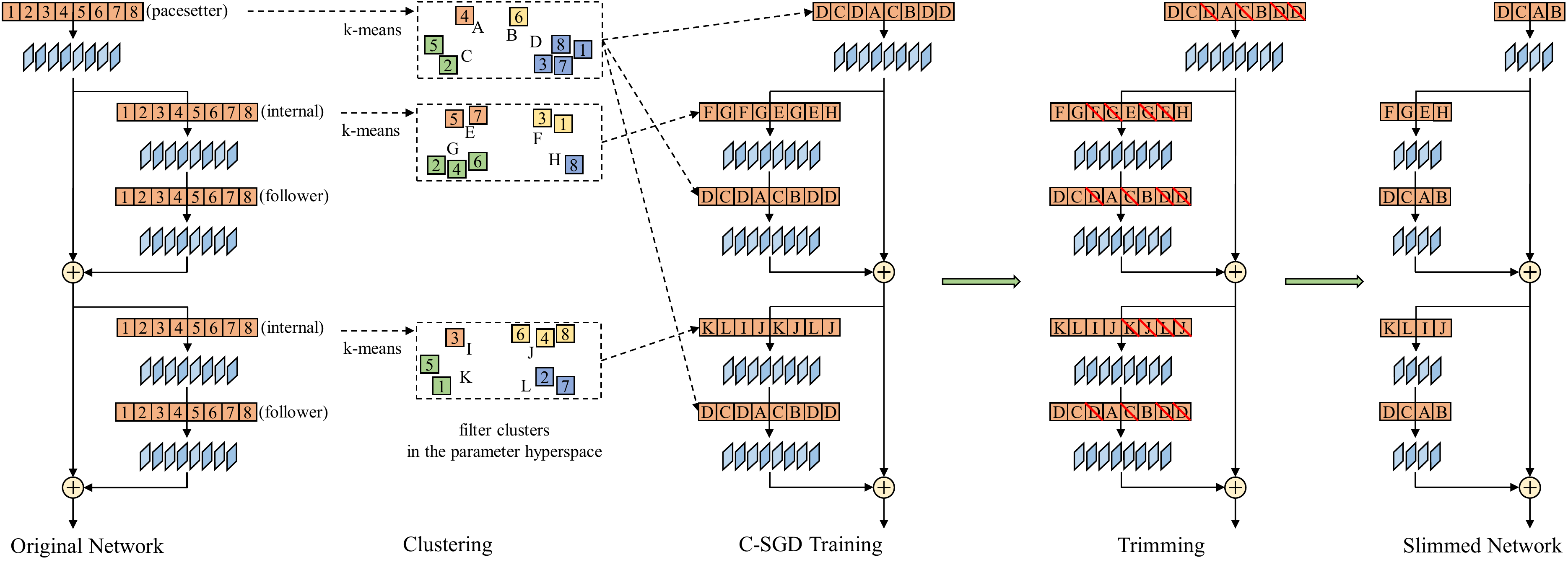}}
		\caption{Sketch for slimming ResNets. We take the first stage of a toy ResNet where every layer has 8 filters for example. Since every convolutional layer is directly followed by exactly one batch normalization layer, we view them as a whole. We generate clusters for the pacesetter and internal layers in each stage by k-means for example. Before C-SGD training, the clustering result of a pacesetter is assigned to its followers in order to produce the same redundancy pattern.}
		\label{res_sketch}
	\end{center}
	\vskip -10pt
\end{figure*}
\begin{figure*}
	\begin{center}
		\setlength{\abovecaptionskip}{5pt} 
		\centerline{\includegraphics[width=\linewidth]{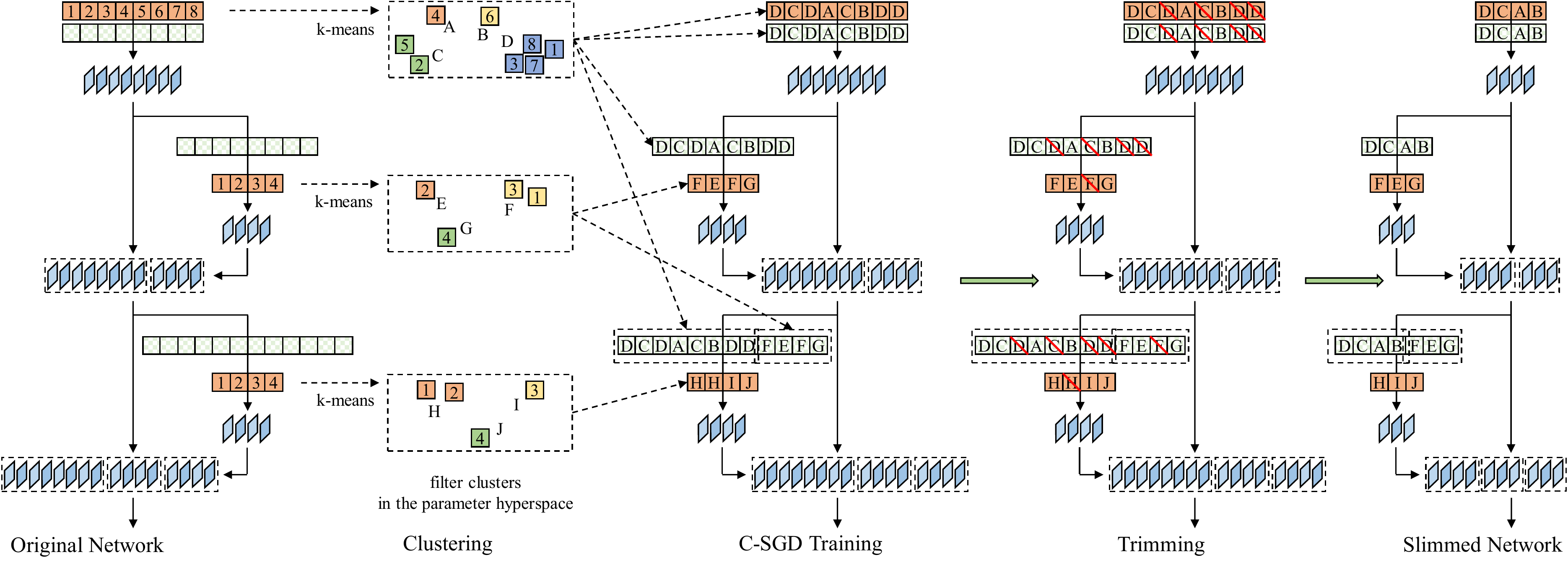}}
		\caption{Sketch for slimming DenseNets. We take a toy DenseNet with growth rate 4 for example. Considering the special dense connection and pre-activation structure of DenseNets, we treat the batch normalization layers separately, which are denoted by the rectangles with chessboard-like background. As the output feature map of every convolutional layer serves as the input of one or more batch normalization layers, we generate clusters for every convolutional layer and apply the clustering results $\mathcal{C}$ to every following batch normalization layer \textit{at the corresponding position}, such that the gradients of $\bm{\gamma}$ and $\bm{\beta}$ are transformed as the preceding convolutional layers. Note that a batch normalization layer can be regarded as a degraded case of the definition in Sect. \ref{sec-formulation} without loss of generality.}
		\label{dense_sketch}
	\end{center}
	\vskip -5pt
\end{figure*}
Recently, accompanied by the advancement of CNN design philosophy, several efficient and compact CNN architectures~\cite{he2016deep,huang2017densely} have emerged and become favored in the real-world applications. Although some excellent works~\cite{kim2015compression,hu2016network,molchanov2016pruning,zhou2016less,yu2018nisp} have shown that the classical plain CNNs, \eg, AlexNet~\cite{krizhevsky2012imagenet} and VGG~\cite{simonyan2014very}, are highly redundant and can be pruned significantly, the pruned versions are usually still inferior to the more up-to-date and complicated CNNs in terms of both accuracy and efficiency.

Filter pruning for very deep and complicated CNNs is challenging due to: \textbf{1)} Firstly, these models are designed in consideration of the computational efficiency, which makes them inherently compact and efficient. \textbf{2)} Secondly, these networks are significantly deeper than the classical ones, thus layer-by-layer pruning becomes too time-consuming, and the errors can increase dramatically when propagated through multiple layers, thus making the estimation of filter importance less accurate~\cite{yu2018nisp}. \textbf{3)} Most importantly, some innovative structures are heavily used in these networks, \eg, shortcuts~\cite{he2016deep} and dense connections~\cite{huang2017densely}, raising an open problem of constrained filter pruning.

For example, in each stage of ResNets, every residual block is expected to add the learned residuals to the stem feature maps produced by the first or the projection layer (referred to as \textit{pacesetter}), thus the last layer of every residual block (referred to as \textit{follower}) must be pruned in the same pattern as the pacesetter, \ie, the remaining set $\mathcal{R}$ of all the followers and the pacesetter must be the same, or the model will be damaged so badly that finetuning cannot restore its accuracy. However, important filters in the pacesetters and followers usually reside in different positions, such that we have to prune some important filters in some layers due to the constraints. An intuitive explanation is shown in Fig. \ref{fig-res-break}.

In some successful explorations, Li \etal~\cite{li2016pruning} sidestep this problem by only pruning the \textit{internal} layers on ResNet-56, \ie, the first layers in the residual blocks. Liu \etal~\cite{liu2017learning} and He \etal~\cite{he2017channel} skip these troublesome layers and insert an extra sampler layer before the internal layers during inference time to reduce the input channels. Though these methods can prune the models to some extent, from a holistic perspective, the networks are not literally ``slimmed'' but actually ``clipped'', as shown in Fig. \ref{fig-slim-clip}.

We have partly solved this open problem by C-SGD, where the key is to force different layers to \textit{learn the same redundancy pattern}. For example, if the layer $p$ and $q$ have to be pruned in the same pattern, we only generate clusters for the layer $p$ by some means and assign the resulting cluster set to the layer $q$, namely, $\mathcal{C}_q\gets\mathcal{C}_p$. Then during C-SGD training, the same redundancy patterns among filters at both layer $p$ and $q$ are produced. \Ie, if the $j$-th and $k$-th filters at layer $p$ become identical, we ensure the sameness of the $j$-th and $k$-th filters at layer $q$ as well, thus the troublesome layers can be pruned along with others with no performance loss. Fig. \ref{res_sketch} and Fig. \ref{dense_sketch} illustrate how we prune ResNets and DenseNets, respectively, where each rectangle represents a filter, and different filters labeled by the same letter become identical during C-SGD training.

\subsection{C-SGD for Scaling and Squeezing: a New Training Methodology}

In this paper, we propose Scaling and Squeezing, a novel CNN training methodology based on C-SGD, to improve the performance of CNN without any extra parameters or FLOPs. Concretely, given an off-the-shelf CNN architecture, we first train a model with regular SGD from scratch, where some layers are wider than the original. Naturally, such a wide model will be more powerful than the original one but at the costs of more parameters and computations. Then we use C-SGD to slim it down to the original structure. As will be shown in Sect. \ref{sect-scale-squeeze}, the performance of the resulting model will be lower than the wide one, but higher than a counterpart with the same structure trained with regular SGD. Intuitively, when the filters in each cluster are constrained to grow closer, the learned knowledge is gradually ``squeezed'' into the cluster center, \ie, the merged filter, such that the resulting model becomes more powerful than the normal counterpart. Interestingly, it is observed that scaling and pruning globally, including those troublesome layers, yields better performance than only scaling and pruning the easy-to-prune layers. This observation highlights the significance of C-SGD in partly solving the constrained filter pruning problem.

The key distinguishing Scaling and Squeezing from simply pruning a bigger model into a smaller one with the traditional pruning methods is that the former improves the performance by a significant margin while the latter cannot outperform a regularly trained counterpart due to the weakness of their non-ideal redundancy patterns. Recently, Liu \etal~\cite{liu2018rethinking} validated several pruning-and-finetuning methods~\cite{han2015learning,molchanov2016pruning,li2016pruning,luo2017thinet,he2017channel,liu2017learning} and empirically found out that with the same width, the network obtained by pruning delivers no better performance than a counterpart trained from scratch. The authors state that though the remaining weights are considered \textit{important} by the pruning criteria, inheriting them does not help the finetuning process achieve better accuracy, but might trap the pruned model into a bad local minimum. However, our method does not judge the parameters by their importance and discard the unimportant ones, nor finetune a model after lossy pruning. On the contrary, by averaging the gradients of filters in each cluster (Eq. \ref{update-rule}), we fully utilize the information encoded in the objective function to supervise the whole cluster and reduce the possibility of being trapped into a local minimum.

Of note is that Scaling and Squeezing manipulates the model's structure for higher accuracy, which is complementary to the other techniques for improving CNN performance like stronger data augmentation, advanced loss functions, etc. Though Scaling and Squeezing increases the training costs, it is still practical as a methodology to boost the performance of CNNs, because we usually care about the inference-time performance and efficiency more than the training costs in real-world applications, as we commonly train our models on the powerful GPU/TPU workstations and deploy them to multiple front-end devices where the efficiency matters. By Scaling and Squeezing, we obtain a model of the same structure as a normally trained counterpart, thus the increased accuracy can be regarded as free benefits, from the viewpoint of the end users.

\section{Experiments}
We performed abundant experiments to evaluate C-SGD. \textbf{1)}~We validated the effectiveness of C-SGD by pruning several common benchmark models on CIFAR-10~\cite{krizhevsky2009learning} and comparing with the reported results from several recent competitors. \textbf{2)} We justified the practicability of C-SGD on real-world applications by pruning on ImageNet~\cite{deng2009imagenet}. \textbf{3)} We validated the generalization performance of C-SGD on COCO detection and VOC segmentation. \textbf{4)} We compared different clustering methods and discovered a minor difference. \textbf{5)} We demonstrated the superiority of C-SGD over the zeroing-out methods in the sense that C-SGD converges faster and enables lossless pruning by producing ideal redundancy patterns. \textbf{6)} A series of controlled experiments were conducted to fairly compare C-SGD and some other pruning methods with the same training settings. \textbf{7)} We found out that when both trained from scratch, a model with identical filters can outperform another one without, thus providing empirical evidence supporting the assumption that redundancy can help the convergence of neural network. \textbf{8)} We justified the significance of solving the constrained filter pruning problem by showing that global slimming on ResNet yields better performance than simply clipping the easy-to-prune layers. \textbf{9)} We verified the effectiveness of Scaling and Squeezing by training a model with the same architecture but wider layers, squeezing it into the original width and comparing with the normally trained counterparts.

\subsection{Pruning Results on CIFAR-10}\label{sect-exp-cifar}
\begin{table*}
	\caption{Pruning Results on CIFAR-10 sorted by the FLOPs reduction ratio. Note that a negative error increase denotes an improvement in the accuracy. For ResNets, ``internal'' and ``sampler'' denote that the architecture is still 16-32-64, but the internal layers of residual blocks are clipped, or the sampler layers are inserted in front of the blocks, as described in Fig. \ref{fig-slim-clip}.}
	\label{exp-table-cifar}
	\begin{center}
				\resizebox{\textwidth}{!}{
		\begin{tabular}{llccccccc}
			\toprule
			Model		&	Result 						&Base Top1	&Pruned Top1						& \makecell{Top1 Error \\ Abs/Rel $\uparrow$\% }	& 	\makecell{FLOPs \\ $\downarrow$\%} & 	\makecell{Params \\ $\downarrow$\%}	&	Architecture		\\
			\midrule	
			VGG			&	Li \etal~\cite{li2016pruning}					&	93.25	&	93.40			&	-0.15 / -2.22	&	34.2	&	64.0	&	-	\\
			VGG			&	Network Slimming~\cite{liu2017learning}&	93.66	&	93.80			&	-0.14 / -2.20	&	51.0	&	88.5	&	-	\\
			VGG			&	Hu \etal~\cite{hu2018novel}						&	92.71	&	92.74			&	-0.03 / -0.41	&	56.2	&	84.0	&	-	\\
			VGG			&	GSFP~\cite{xu2018globally}				&	93.25	&	93.29			&	-0.04 / 0.59	&	61.46	&	74.21	&	-	\\
			\textbf{VGG}&	\textbf{C-SGD-VGG-A}		&	\textbf{93.53}	&	\textbf{94.10}		&	\textbf{-0.57 / -8.80}	&	\textbf{61.69}	&\textbf{86.28}	&	-	\\	
			VGG			&	Jiang \etal~\cite{jiang2018efficient}				&	93.46	&	93.40			&	0.06 / 0.91		&	67.6	&	92.7	&	-	\\
			VGG			&	Zhu \etal~\cite{zhu2018improving}				&	93.58	&	93.31			&	0.27 / 4.20		&	68.75	&	88.23	&	-	\\
			VGG			&	Zhou \etal~\cite{zhou2018network}					&	91.0	&	90.6			&	0.4 / 4.44		&	71.42	&	97.44	&	-	\\		
			VGG			&	2PFPCE~\cite{min20182pfpce}			&	92.98	&	92.76			&	0.22 / 3.13		&	74.83	&	74.53	&	-	\\
			\textbf{VGG}&	\textbf{C-SGD-VGG-B}		&	\textbf{93.53}	&	\textbf{93.78}		&	\textbf{-0.25 / -3.86}	&	\textbf{75.15}	&\textbf{90.09}	&	-	\\	
			VGG			&	Huang \etal~\cite{huang2018learning}				&	92.77	&	89.37			&	3.40 / 47.02	&	80.6	&	92.8	&	-	\\
			VGG			&	Ding \etal~\cite{ding2018auto}					&	92.92	&	92.44			&	0.48 / 6.77		&	81.39	&	93.51	&	-	\\
			VGG			&	Singh \etal~\cite{singh2018stability}				&	93.49	&	93.02			&	0.47 / 7.21		&	83.43	&	95.83	&	-	\\
			\textbf{VGG}&	\textbf{C-SGD-VGG-C	}		&	\textbf{93.53}	&	\textbf{93.59}		&	\textbf{-0.06 / -0.92}	&	\textbf{85.02}	&\textbf{96.54}	&	-	\\	
			\textbf{VGG}&	\textbf{C-SGD-VGG-D}		&	\textbf{93.53}	&	\textbf{92.20}		&	\textbf{1.33 / 20.55}	&	\textbf{90.12}	&\textbf{97.95}	&	-	\\	
			\midrule
			Res56	&	Li \etal~\cite{li2016pruning}					&	93.04	&	93.06			&	-0.02 / -0.28	&	27.60	&	13.7	&	internal \\
			Res56	&	NISP-56~\cite{yu2018nisp}				&	-		&	-				&	0.03 / -		&	43.61	&	42.60	&	-	\\
			Res56	&	Zhu \etal~\cite{zhu2018improving}				&	93.39	&	93.40			&	-0.01 / -0.15	&	47.36	&	52.38	&	-	\\
			Res56	&	Channel Pruning~\cite{he2017channel}	&	92.8	&	91.8			&	1.0 / 13.88		&	50		&	-		&	sampler	\\
			Res56	&	ADC~\cite{he2018adc}					&	92.8	&	91.9			&	0.9	/ 12.5		&	50		&	-		&	sampler	\\
			Res56	&	FPGM~\cite{FPGM}						&	93.59	&	93.26			&	0.33 / 5.14		&	52.6	&	-		&	-		\\
			Res56	&	LFPC~\cite{he2020learning}				&	93.59	&	93.24			&	0.35 / 5.46		&	52.9	&	-		&	-		\\
			Res56	&	AFP~\cite{ding2018auto}				&	93.93	&	92.94			&	0.99 / 16.30	&	60.85	&	60.90	&	10-20-40	\\
			\textbf{Res56}	&	\textbf{C-SGD-Res56-10-20-40}&\textbf{93.39}	&\textbf{93.62}			&	\textbf{-0.23 / -3.47}	&	\textbf{60.85}	&	\textbf{60.90}	&	\textbf{10-20-40}	\\
			\midrule
			Res110	&	Li \etal~\cite{li2016pruning}					&	93.53	&	93.30			&	0.23 / 3.55		&	38.60	&	32.4	&	internal \\
			Res110	&	NISP-110~\cite{yu2018nisp}				&	-		&	-				&	0.18 / -		&	43.78	&	43.25	&	-	\\
			Res110	&	GAL-0.5~\cite{GAL}									&	93.50		&	92.74			&	0.76 / 11.6		&	48.5	&	-	&	-	\\
			Res110	&	HRank~\cite{HRank}						&	93.50		&	93.36			&	0.14 / 2.15		&	58.2	&	-	&	-	\\
			\textbf{Res110}	&	\textbf{C-SGD-Res110-10-20-40}&	\textbf{94.38}	&	\textbf{94.41}	&	\textbf{-0.03 / -0.53} 				&	\textbf{60.89}	& \textbf{60.92}	&	\textbf{10-20-40}	\\
			\midrule
			Res164	&	Network Slimming~\cite{liu2017learning}&	94.58	&	94.73			&	-0.15 / -2.76	&	44.90	&	35.2	&	sampler	\\
			\textbf{Res164}	&	\textbf{C-SGD-Res164-12-24-46}		&	\textbf{94.83}	&	\textbf{95.08}	&	\textbf{-0.25 / -4.83}	&	\textbf{45.24}	&	\textbf{54.75}	&	\textbf{12-24-46}	\\
			\textbf{Res164}	&	\textbf{C-SGD-Res164-10-20-40}		&	\textbf{94.83}	&	\textbf{94.81}	&	\textbf{0.02 / 0.38}	&	\textbf{60.91}	&	\textbf{60.93}	&	\textbf{10-20-40}	\\	
			\midrule	
			Dense40	&	Network Slimming~\cite{liu2017learning}&	93.89	&	94.35			&	-0.46 / -7.52	&	55.00	&	65.2	&	-	\\
			\textbf{Dense40}&	\textbf{C-SGD-Dense40-5-8-10}	&	\textbf{93.81}	&	\textbf{94.56}		&	\textbf{-0.75 / 12.11}&	\textbf{60.05}	& \textbf{36.16}	&	\textbf{5-8-10}\\			
			\bottomrule
		\end{tabular}
				}
	\end{center}
\end{table*}

In this subsection, we aim to preliminarily evaluate C-SGD by some pruning experiments and comparisons with the recent competitors on CIFAR-10 (Table. \ref{exp-table-cifar}). Our results are marked by bold font. Since our base models deliver different accuracy than the competitors, we present the absolute and relative error increase, which are commonly adopted as the metrics to compare the change of accuracy on different base models. For example, the Top1 accuracy is 93.53\% for our base VGG~\cite{simonyan2014very} model and 92.20\% for the result labeled as C-SGD-VGG-D, such that the absolute and relative error increase are $93.53\%-92.20\%=1.33\%$ and $\frac{1.33}{100-93.53}=20.55\%$, respectively. For each trial we start from a well-trained base model, cluster the filters by k-means, apply C-SGD training on all the layers \textit{simultaneously}, prune every layer and test the resulting model.

The base models are trained from scratch for 600 epochs to ensure the convergence, which is longer than the usually adopted benchmarks (160~\cite{he2016deep} or 300~\cite{huang2017densely} epochs), because we expect to perform pruning on a fully trained base model, such that the accuracy increase (in the case of VGG, Res56, Res110, Dense40) cannot be simply attributed to the training on a base model which has not fully converged. We use the data augmentation techniques adopted by He \etal~\cite{he2016deep}, \ie, padding to $40\times40$, random cropping and flipping. The hyper-parameter $\epsilon$ is casually set to $3\times10^{-3}$. We perform C-SGD training for 600 epochs with batch size 64 and a learning rate initialized as $3\times10^{-2}$ then multiplied by 0.1 when the loss stops decreasing. 

We start with VGG, a 13-layer plain network. As a common practice, every convolutional layer is followed by a batch normalization. In order to compare with other competitors with different pruning ratios, we prune the base model with four different target width settings labeled from A to D to reach around 60\%, 75\%, 85\%, 90\% FLOPs reduction, respectively, which are shown in Table. \ref{exp-table-vgg-width}. As we do not intend to push the state-of-the-art results on such a simple network, we set the target width casually. \Eg, the 13 layers of the model labeled as C-SGD-VGG-C has 20, 50, 80, 80, 80, 80, 80, 60, 60, 60, 60, 60, 60 filters, respectively, such that around 85\% FLOPs are reduced. To obtain a series of models with descending FLOPs, we iteratively apply C-SGD pruning, \eg, C-SGD-VGG-C is the input to C-SGD-VGG-D. The comparison shows the superiority of C-SGD over the competitors. Though some prior works succeeded in pruning VGG with improved performance, they did not achieve an increase so significant at such high pruning ratios (\eg, 0.25\% at 75.15\%). Moreover, we observe no accuracy drop even when the FLOPs reduction reaches 85.15\%. Of note is that the comparisons are biased towards the competitors as it is more challenging to prune a higher-accuracy model without accuracy drop.

For ResNets, we aim to reduce around 60\% FLOPs of every model by pruning $3/8$ of \textit{every} convolutional layer, thus the parameters and FLOPs are reduced by $1-(5/8)^2=61\%$. As the original ResNets have 16, 32 and 64 filters at each layer in the three stages, respectively, we denote their structure as 16-32-64, and our pruned models as 10-20-40. Aggressive as the pruning is, we observe no obvious accuracy drop. Better still, the increased depth does not degrade the effectiveness of C-SGD, which distinguishes C-SGD from the layer-by-layer methods. We also produced a ResNet-164 labeled as 12-24-46, such that its FLOPs are comparable with Liu \etal~\cite{liu2017learning}.

The original DenseNet-40 has 12 filters at every incremental convolutional layer, but the pruned model has 5, 8 and 10 filters for the three stages, respectively, such that the FLOPs are reduced by 60.05\%, and an accuracy increase is achieved, which is consistent with but better than that of Liu \etal~\cite{liu2017learning}.

\begin{table}[t]
	\caption{Width settings of four pruned VGG models labeled from A to D.}
	\label{exp-table-vgg-width}
	\begin{center}
		\begin{small}
			\begin{tabular}{lccccc}
				\toprule
				Layer	   	& original 	&A	&B	&C	&D	\\
				\midrule
				1	&64		&20		&20		&20		&20\\
				2	&64		&50		&50		&50		&50\\
				3	&128	&100	&100	&80		&60\\
				4	&128	&120	&120	&80		&60\\
				5 - 7	&256	&200	&150	&80		&50\\
				8	&512	&200	&150	&60		&50\\
				9 - 13	&512	&100	&100	&60		&50\\
				
				\bottomrule
			\end{tabular}
		\end{small}
	\end{center}
\end{table}

\subsection{Pruning Results on ImageNet}
\begin{table*}
	\caption{Pruning ResNet-50 on ImageNet using the same data preprocessing as ThiNet~\cite{luo2017thinet}.}
	\label{exp-table-imagenet-thinet}
	\begin{center}
		\resizebox{\textwidth}{!}{
			\begin{tabular}{lcccccccc}
				\toprule
				Result 							&Base Top1	&Base Top5	&Pruned Top1	&Pruned Top5	& \makecell{Top1 Error \\ Abs/Rel $\uparrow$\% }	&	\makecell{Top5 Error \\ Abs/Rel $\uparrow$\% }	& 	\makecell{FLOPs \\ $\downarrow$\%} & \makecell{Params \\ $\downarrow$\%}				\\
				\midrule
				ThiNet-70						&72.88		&91.14		&72.04			&90.67			&0.84 / 3.09	&0.47 / 5.30	&36.75		&33.72	\\
				\textbf{C-SGD-Res50-70}		&\textbf{74.17}		&\textbf{91.77}		&\textbf{73.80}		&\textbf{91.71}		&\textbf{0.37 / 1.43}	&\textbf{0.06 / 0.72}	&\textbf{36.75}	&\textbf{33.72}\\
				ThiNet-50						&72.88		&91.14		&71.01			&90.02			&1.87 / 6.89	&1.12 / 12.64	&55.76		&51.56	\\
				\textbf{C-SGD-Res50-50}		&\textbf{74.17}		&\textbf{91.77}		&\textbf{73.00}		&\textbf{91.42}		&\textbf{1.17 / 4.52}	&\textbf{0.35 / 4.25}	&\textbf{55.76}	&\textbf{51.56}\\
				\bottomrule
			\end{tabular}
		}
	\end{center}
\end{table*}
\begin{table*}
	\caption{Pruning ResNet-50 on ImageNet using bounding box distortions and color shift, sorted by the FLOPs reduction ratio.}
	\label{exp-table-standard}
	\begin{center}
		\resizebox{\textwidth}{!}{
			\begin{tabular}{lcccccccc}
				\toprule
				Result 							&Base Top1	&Base Top5	&Pruned Top1	&Pruned Top5	& \makecell{Top1 Error \\ Abs/Rel $\uparrow$\% }	&	\makecell{Top5 Error \\ Abs/Rel $\uparrow$\% }	& 	\makecell{FLOPs \\ $\downarrow$\%}	& \makecell{Params \\ $\downarrow$\%} 			\\
				\midrule
				\textbf{C-SGD-Res50-70}	&	\textbf{75.33}	&	\textbf{92.56}	&	\textbf{75.27}		&	\textbf{92.46}		&	\textbf{0.06 / 0.24}  	&	\textbf{0.10 / 1.34}	&	\textbf{36.75}&\textbf{33.38}\\	
				NISP~\cite{yu2018nisp}			&	-		&	-		&	-			&	-			&	0.89 / -		&	- / -			&	44.01	&	43.82	\\	
				Singh \etal~\cite{singh2018stability}		&	-		&	92.65	&	-			&	92.2		&	- / -			&	0.45 / 6.13		&	44.45	&	40.92	\\	
				\textbf{C-SGD-Res50-60}		&	\textbf{75.33}	&	\textbf{92.56}	&	\textbf{74.93}		&	\textbf{92.27}		&	\textbf{0.40 / 1.62}	&	\textbf{0.29 / 3.89}	&	\textbf{46.24}&	\textbf{42.83}	\\
				CFP~\cite{singh2018leveraging}	&	75.3	&	92.2	&	73.4		&	91.4		&	1.9 / 7.69		&	0.8 / 10.25		&	49.6	&	-	\\
				Channel Pr~\cite{he2017channel}	&	- 		&	92.2	&	-			&	90.8		&	- / -			&	1.4 / 17.94	&	50		&	-	\\
				SPP~\cite{wang2017structured}	&	- 		&	91.2	&	-			&	90.4		&	- / -			&	0.8 / 9.09		&	50		&	-	\\
				HP~\cite{xu2018hybrid}			&	76.01 	&	92.93	&	74.87		&	92.43		&	1.14 / 4.75		&	0.50 / 7.07		&	50		&	32.5\\
				ELR~\cite{wang2018exploring}	&	- 		&	92.2	&	-			&	91.2		&	- / -			&	1 / 12.82		&	50		&	-	\\
				GDP~\cite{lin2018accelerating}	&	75.13 	&	92.30	&	71.89		&	90.71		&	3.24 / 13.02	&	1.59 / 20.64	&	51.30	&	-	\\
				SSR-L2~\cite{lin2019towards}	&	75.12 	&	92.30	&	71.47		&	90.19		&	3.65 / 14.67	&	2.11 / 27.40	&	55.76	&	51.56\\	
				\textbf{C-SGD-Res50-50}				&	\textbf{75.33}	&	\textbf{92.56}	&	\textbf{74.54}		&	\textbf{92.09}		&	\textbf{0.79 / 3.20}	&	\textbf{0.47 / 6.31}	&	\textbf{55.76}&\textbf{51.50}\\		
				ThiNet~\cite{DBLP:journals/pami/LuoZZXWL19} & 75.30 & 92.20	&72.03 &90.99				&	3.27 / 13.23		&	1.21 / 15.51&	55.83	&	-	\\
				\bottomrule
			\end{tabular}
		}
	\end{center}
\end{table*}
\begin{table*}
	\caption{Pruning the standard torchvision ResNet-50 (denoted by Res50B) on ImageNet using default data augmentation.}
	\label{exp-table-torchvision}
	\begin{center}
		\resizebox{\textwidth}{!}{
			\begin{tabular}{lcccccccc}
				\toprule
				Result 							&Base Top1	&Base Top5	&Pruned Top1	&Pruned Top5	& \makecell{Top1 Error \\ Abs/Rel $\uparrow$\% }	&	\makecell{Top5 Error \\ Abs/Rel $\uparrow$\% }	& 	\makecell{FLOPs \\ $\downarrow$\%}	& \makecell{Params \\ $\downarrow$\%} 			\\
				\midrule
				\textbf{C-SGD-Res50B-70}	&	\textbf{76.15}	&	\textbf{92.87}	&	\textbf{75.94}		&	\textbf{92.88}		&	\textbf{0.21 / 0.88}  	&	\textbf{-0.01 / -0.14}	&	\textbf{36.38}&\textbf{33.38}\\	
				SFP~\cite{he2018soft}			&	76.15 	&	92.87	&	74.61		&	92.06		&	1.54 / 6.45		&	0.81 / 11.36	&	41.8	&	-	\\
				GAL-0.5~\cite{GAL}				&	76.15	&	92.87	&	71.95		&	90.94		&	4.20 / 17.61	&	1.93 / 27.06	&	43.03	&	-	\\
				HRank~\cite{HRank}				&	76.15	&	92.87	&	74.98		&	92.33		&	1.17 / 4.90		&	0.54 / 7.57		&	43.76	&	-	\\
				\textbf{C-SGD-Res50B-60}	&	\textbf{76.15}	&	\textbf{92.87}	&	\textbf{75.80}		&	\textbf{92.65}		&	\textbf{0.35 / 1.46}  	&	\textbf{0.22 / 3.08}	&	\textbf{46.51}&\textbf{42.83}\\	
				Autopr~\cite{luo2018autopruner}&	76.15 	&	92.87	&	74.76		&	92.15		&	1.39 / 5.82		&	0.72 / 10.09	&	51.21	&	-	\\
				FPGM~\cite{FPGM}				&	76.15	&	92.87	&	74.83		&	92.32		&	1.32 / 5.53		&	0.55 / 7.71		&	53.5	&	-	\\
				\textbf{C-SGD-Res50B-50}	&	\textbf{76.15}	&	\textbf{92.87}	&	\textbf{75.29}		&	\textbf{92.39}		&	\textbf{0.86 / 3.60}  	&	\textbf{0.48 / 6.73}	&	\textbf{55.44}&\textbf{51.50}\\	
				\bottomrule
			\end{tabular}
		}
	\end{center}
\end{table*}

Table. \ref{exp-table-imagenet-thinet} and Table. \ref{exp-table-standard} show the pruning results of our method and other works on the original ResNet-50 \cite{he2016deep}, which is a commonly adopted benchmark in the filter pruning literature. Since many competitors experimented with the torchvision \cite{torch-model} version of ResNet-50 (denoted by Res50B), we also prune it for the fair comparison (Table. \ref{exp-table-torchvision}). The only difference between the original ResNet-50 and Res50B is that the former conducts downsampling by the $1\times1$ conv at the beginning of a stage while the latter uses the $3\times3$ conv. For pruning each model, we train with C-SGD for 70 epochs with a learning rate initialized as 0.03 and multiplied by 0.1 at the 30th, 50th and 60th epochs, respectively. We use a batch size of 256 on 8 GPUs, weight decay of $10^{-4}$, centripetal strength $\epsilon=0.05$.

First, we compare C-SGD with ThiNet~\cite{luo2017thinet,DBLP:journals/pami/LuoZZXWL19}, a classic filter pruning method (Table. \ref{exp-table-imagenet-thinet}). For the fair comparison, we augment the training data in the same way~\cite{luo2017thinet}, \ie, the images are simply resized to $256\times256$, then $224\times224$ random cropping and horizontal flipping are adopted. Because of such weak image distortion, the accuracy of our base model is lower than that reported in the ResNet paper~\cite{he2016deep}. At test time, we use a single central crop. Following ThiNet, all the internal layers in each residual block of C-SGD-Res50-70 and C-SGD-Res50-50 are shrunk to 70\% and 50\% of the original width, respectively.

Then, we provide the comparison of C-SGD and some more recent competitors using the standard data augmentation methods including bounding box distortions and color shift (Table. \ref{exp-table-standard}). Our base model reaches a Top1/Top5 accuracy of 75.33\%/92.56\%. Though the base models of some competitors have different accuracies, the results are still comparable in terms of the absolute and relative error increase. Following ThiNet and Lin \etal~\cite{lin2019towards}, we slim the internal layers down to 70\%, 60\% and 50\% of the original width, respectively.

With Res50B, we use the official pre-trained model and the default official data preprocessing \cite{torch-example} (Table. \ref{exp-table-torchvision}). 

As can be observed, our pruned models exhibit fewer FLOPs and lower error increase. Of note is that, instead of carefully tuning the target network width, we simply apply the same pruning ratio for each internal layer. Better results are promising to be achieved if more layer sensitivity analyzing experiments~\cite{li2016pruning,he2017channel,yu2018nisp} are conducted, and the target network structures are tuned accordingly.

Another set of experiments are conducted on DenseNet-121~\cite{huang2017densely} (Table. \ref{exp-table-dense121}). Without consideration of the layers' sensitivity or the filters' importance, the same pruning ratio is applied for each stage. For C-SGD-Dense121-A, the internal layers, \ie, the first layers in each dense block, are shrunk to 7/8 of the original width, and the incremental factors of the first three stages become 18, 20, 24, respectively. For C-SGD-Dense121-B, the internal layers are further slimmed down to 3/4 of the original width. We prune the lower-level layers harder than the higher-level ones not because of any prior knowledge about the model, but simply because such layers operate on higher-resolution feature maps so that reducing their width results in higher acceleration. Though DenseNet-121 is not usually chosen as a benchmark model for pruning because of its complicated and compact structure, we can slim it with minor decrease in the accuracy. Such a success shows the significance of C-SGD in solving the constrained filter pruning problem and the effectiveness of the strategy described in Fig. \ref{dense_sketch}.

\begin{table*}
	\caption{Pruning DenseNet-121 on ImageNet.}
	\label{exp-table-dense121}
	\begin{center}
		\resizebox{\textwidth}{!}{
			\begin{tabular}{lcccccccc}
				\toprule
				Result 							&Base Top1	&Base Top5	&Pruned Top1	&Pruned Top5	& \makecell{Top1 Error \\ Abs/Rel $\uparrow$\% }	&	\makecell{Top5 Error \\ Abs/Rel $\uparrow$\% }	& 	\makecell{FLOPs \\ $\downarrow$\%} 	& \makecell{Params \\ $\downarrow$\%} 			\\
				\midrule
				C-SGD-Dense121-A 			&	74.47 	&	92.14	&	74.25		&	91.76		&	0.22 / 0.86		&	0.38 / 4.83		&	34.65	&21.53\\
				C-SGD-Dense121-B			&	74.47	&	92.14	&	73.73		&	91.55		&	0.74 / 2.89 	&	0.59 / 7.50		&	42.28	&29.89\\
				\bottomrule
			\end{tabular}
		}
	\end{center}
\end{table*}

\subsection{Semantic Segmentation and Object Detection}

We verify the effectiveness of C-SGD on the downstream tasks including semantic segmentation and object detection. 

First, we use the augmented VOC 2012 dataset for semantic segmentation as a common practice \cite{DBLP:conf/iccv/HariharanABMM11,DBLP:journals/pami/ChenPKMY18}, which has 10,582 images for training (\textit{trainaug} set) and 1449 images for validation (\textit{val} set). We construct a PSPNet \cite{DBLP:conf/cvpr/ZhaoSQWJ17} with the original pre-trained ResNet-50B as the backbone and finetune with a poly learning rate policy with base of 0.01 and power of 0.9, weight decay of $10^{-4}$ and a global batch size of 16 on 4 GPUs for 50 epochs. Then we use the pruned models denoted as C-SGD-Res50B-70 and C-SGD-Res50B-60 (Table. \ref{exp-table-torchvision}) as the backbones and finetune with the identical settings.

Then we experiment with COCO detection. More specifically, the training set is \textit{COCO2017train} and the validation set is \textit{COCO2017val}. We construct a Faster RCNN \cite{ren2015faster} with FPN \cite{lin2017feature} and the original pre-trained ResNet-50B as the backbone. We finetune for 12 epochs with a learning rate initialized as 0.02 and multiplied by 0.1 at the 8th and 11th epochs respectively. Then we use C-SGD-Res50B-70 and C-SGD-Res50B-60 as the backbones again all with the identical settings.

Table. \ref{exp-table-detect} demonstrates the generalization performance of the pruned models, which show very minor or even no decrease in the mIoU on VOC and AP on COCO.

\begin{table}[t]
	\caption{Semantic segmentation results on VOC2012 and object detection results on COCO with the original and pruned ResNet-50B backbones.}
	\label{exp-table-detect}
	\begin{center}
		\begin{small}
			\begin{tabular}{lccccc}
				\toprule
				Backbone	&  \makecell{Top-1 acc on \\ ImageNet} 	&	\makecell{mIoU on \\ VOC}	& \makecell{AP on \\ COCO}		\\
				\midrule
				Original ResNet-50B			&76.15		&	76.29	&	33.2	\\
				C-SGD-Res50B-70	&75.94		&	76.36	&	32.8	\\
				C-SGD-Res50B-60	&75.80		&	75.78	&	33.1	\\
				\bottomrule
			\end{tabular}
		\end{small}
	\end{center}
\end{table}

\subsection{Studies on the Clustering Methods}\label{sec-exp-clustering-methods}
\begin{table*}
	\caption{Pruning results with k-means, even, or imbalanced clustering. }
	\label{exp-table-clustering}
	\begin{center}
		\resizebox{\textwidth}{!}{
			\begin{tabular}{lllccccc}
				\toprule
				Dataset		&	Model		&	Result 				&Base Top1	&\makecell{Pruned Top1 \\ K-means Clustering } &\makecell{Pruned Top1 \\ Even Clustering } &\makecell{Pruned Top1 \\ Imbalanced Clustering } \\
				\midrule
				CIFAR-10	&	VGG			&	C-SGD-VGG-C					&	93.53	&	\textbf{93.59}	&	93.25			&93.19	\\
				CIFAR-10	&	ResNet-56	&	C-SGD-Res56-10-20-40		&	93.39	&	\textbf{93.62}			&	93.44			&93.45		\\
				CIFAR-10	&	ResNet-110	&	C-SGD-Res110-10-20-40			&	94.38	&	94.41			&	\textbf{94.54}	&94.11	\\	
				CIFAR-10	&	ResNet-164	&	C-SGD-Res164-10-20-40			&	94.83	&	\textbf{94.81}	&	94.80			&94.70	\\	
				CIFAR-10	&	DenseNet-40	&	C-SGD-Dense40-5-8-10		&	93.81	&	\textbf{94.56}	&	94.37			&93.94	\\			
				ImageNet	&	ResNet-50	&	C-SGD-Res50-70				&	75.33	&	\textbf{75.27}	&	75.14			&74.93	\\
				\bottomrule
			\end{tabular}
		}
	\end{center}
\end{table*}

To study the effects of different clustering methods, we perform another set of experiments using the same settings as before except for even or imbalanced clustering. As shown in Table. \ref{exp-table-clustering}, k-means outperforms the other two clustering methods by a narrow margin on ImageNet and wins 4 out of 5 cases on CIFAR-10, due to the lower intra-cluster distance in the parameter hyperspace. Interestingly, our experiments indicate that the effectiveness of C-SGD-based pruning does not significantly depend on the quality of filter clusters $\mathcal{C}$, since reasonable performance can be achieved with arbitrarily generated clusters.

\subsection{Making Filters Identical \vs Zeroing Filters Out}\label{sec-vs-zero-out}
As making filters identical and zeroing filters out~\cite{liu2015sparse,alvarez2016learning,wen2016learning,zhou2016less,ding2018auto,lin2019towards} are two means of producing redundancy patterns for filter pruning, we perform controlled experiments on ResNet-56 to investigate the difference. For the fair comparison, we aim to produce the same number of redundant filters in both the network trained with C-SGD and the one with group-Lasso Regularization~\cite{roth2008group}. For C-SGD, the number of clusters at each layer is 5/8 of the number of filters. For Lasso, 3/8 of the original filters in the pacesetters and the internal layers are regularized by group-Lasso, and the followers are handled in the same pattern. We use the aforementioned sum of squared kernel deviation $\chi$ (Eq. \ref{eq-def-chi}) and the \textit{sum of squared kernel residuals} $\phi$ as follows to measure the redundancy, respectively. Let $\mathcal{L}$ be the layer index set and $\mathcal{P}_i$ be the to-be-pruned filter set of layer $i$, \ie, the set of the 3/8 filters with group-Lasso regularization,
\begin{equation}
\quad\quad \phi=\sum_{i\in\mathcal{L}}\sum_{j\in \mathcal{P}_i}\Vert\bm{K}^{(i)}_{:,:,:,j}\Vert_2^2 \,.
\end{equation}

\begin{figure}[t]
	\centering
	\subfloat[Values of $\chi$ or $\phi$.]
	{
		\includegraphics[width=0.48\linewidth]{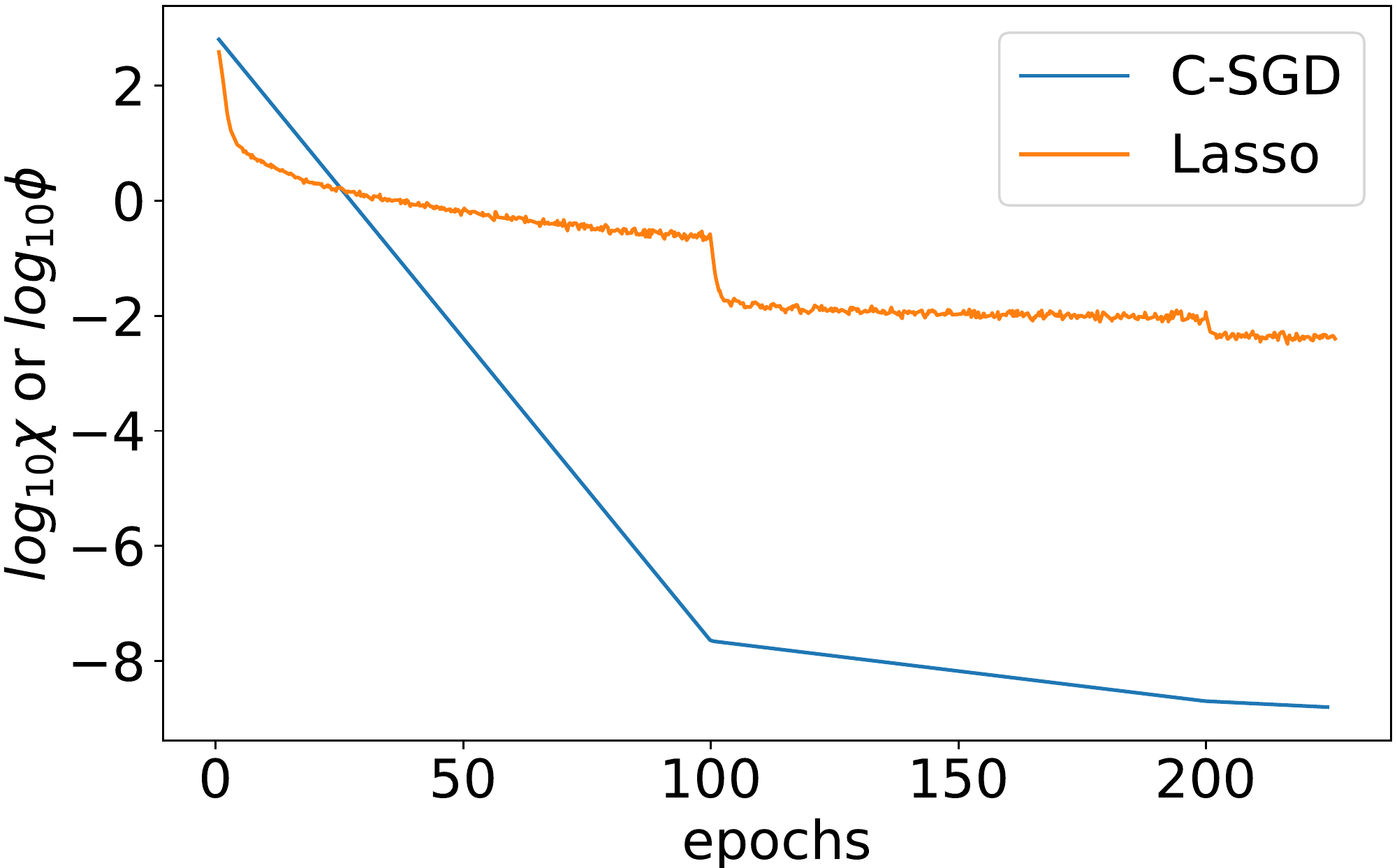}
		\label{curve-chi-phi}
	}
	\subfloat[Validation accuracy.]
	{
		\includegraphics[width=0.48\linewidth]{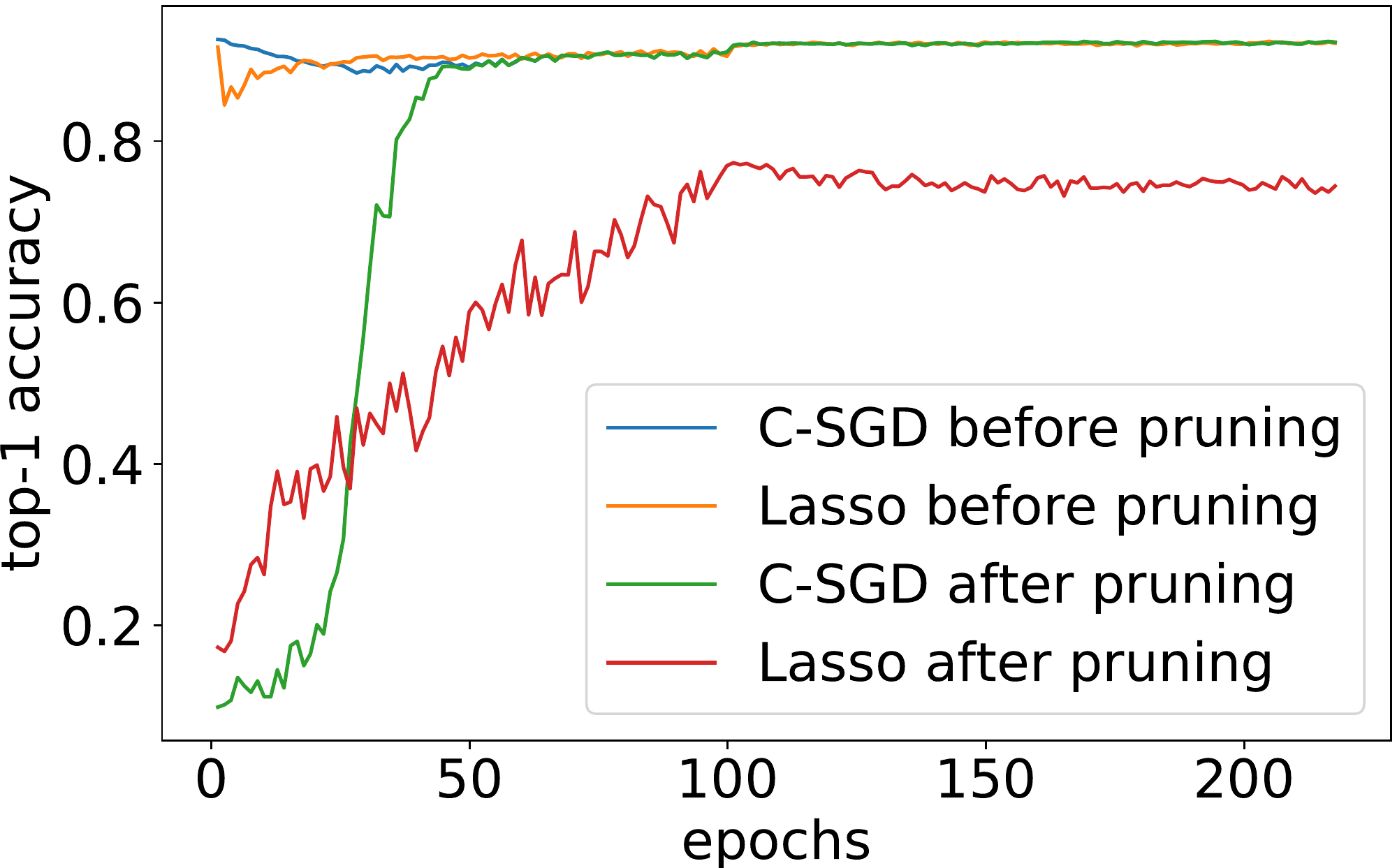}
		\label{curve-acc-epochs}
	}
	\caption{Training process with C-SGD or group-Lasso on ResNet-56. Note the logarithmic scale of the upper figure.}
	\label{fig-chi-phi-acc-before-and-after}
\end{figure}

We present in Fig. \ref{fig-chi-phi-acc-before-and-after} the curves of $\chi$, $\phi$ as well as the validation accuracy both before and after pruning. The learning rate $\tau$ is initially set to $3\times10^{-2}$ and decayed by 0.1 at epoch 100 and 200, respectively. It can be observed that: \textbf{1)} Group Lasso cannot literally zero out filters, but can decrease their magnitude to some extent, as $\phi$ plateaus when the gradients derived from the regularization term become close to those derived from the original objective function. We empirically find out that even when $\phi$ reaches around $4\times10^{-4}$, which is nearly $2\times10^{6}$ times smaller than the initial value, pruning still causes obvious damage (around 10\% accuracy drop). When the learning rate is decayed and $\phi$ is reduced at epoch 200, we observe no improvement in the pruned accuracy, therefore no more experiments with smaller learning rate or stronger group-Lasso Regularization are conducted. We reckon this is due to the error propagation and accumulation in very deep CNNs~\cite{yu2018nisp}. \textbf{2)} By C-SGD, $\chi$ is reduced \textit{monotonically} and perfectly \textit{exponentially}, which suggests faster convergence. In other words, the filters in each cluster can become \textit{infinitely close} to each other at a \textit{constant rate} with a constant learning rate. In the early stage of training, the filters have not become close enough such that pruning degrades the performance (seen from the difference between ``C-SGD before pruning'' and ``C-SGD after pruning'' during the beginning 100 epochs). But after 100 epochs, the pruning causes \textit{absolutely no} damage. \textbf{3)} Training with group-Lasso is $2\times$ slower than C-SGD due to its computational intensity, as implemented using Tensorflow~\cite{abadi2016tensorflow} on GTX 1080Ti GPUs.

\subsection{C-SGD \vs Other Filter Pruning Methods}\label{sec-exp3}
\begin{figure}[t]\label{fig-compare-prune}
	\centering
	\subfloat[Three filters per layer.]
	{
		\includegraphics[width=0.97\linewidth]{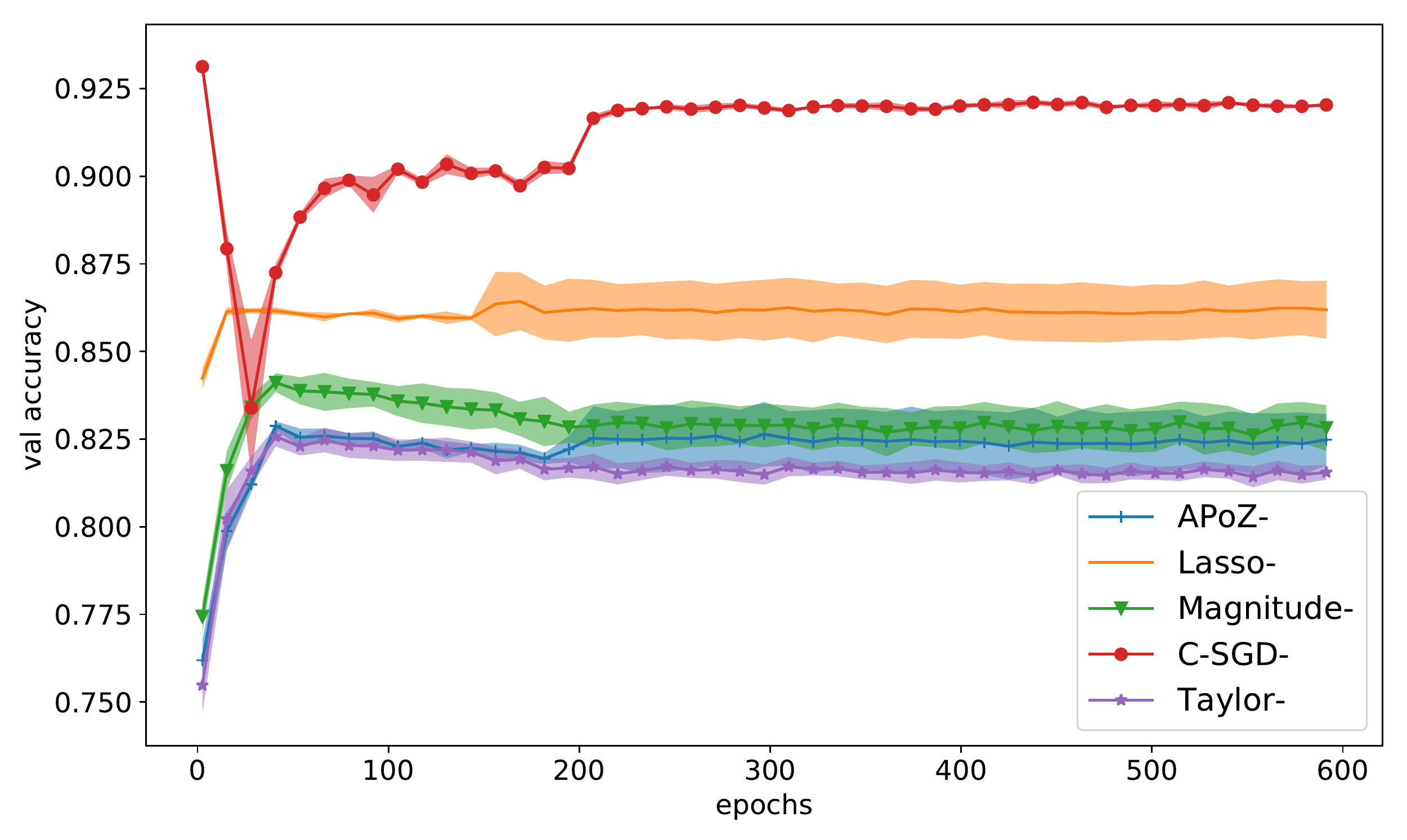}
		\label{densenet40-3}
	}
	\hspace{0.05\linewidth}
	\subfloat[Six filters per layer.]
	{
		\includegraphics[width=0.97\linewidth]{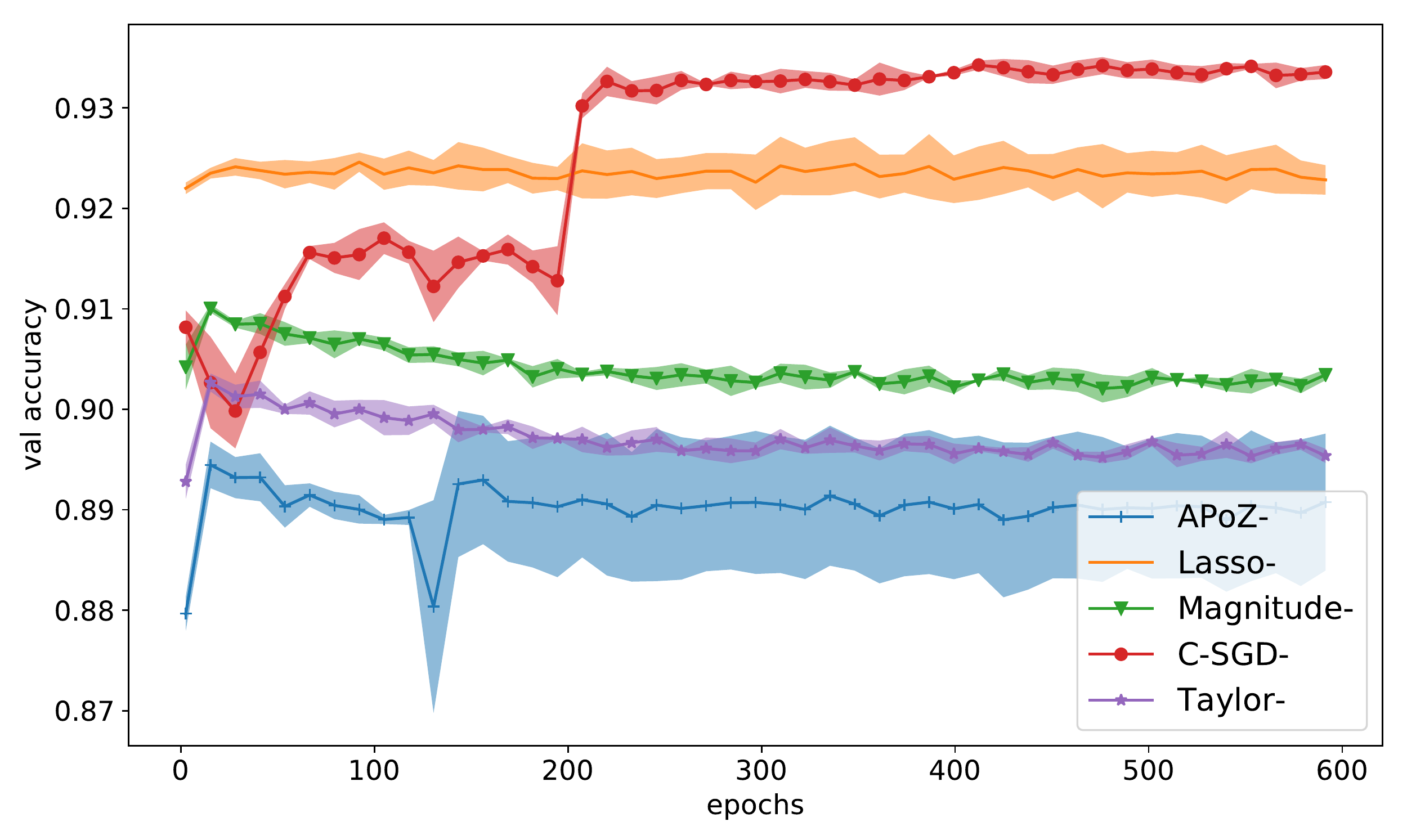}
		\label{densenet40-6}
	}
	\caption{Controlled pruning experiments on DenseNet-40.}
	\label{pruning-dc40}
\end{figure}
In this subsection, we compare C-SGD with other filter pruning methods through a series of controlled experiments on DenseNet-40~\cite{huang2017densely}. We slim \textit{every} incremental convolutional layer of a well-trained DenseNet-40 to 3 and 6 filters, respectively. The experiments are repeated 3 times, and the mean $\pm$ std curves are presented in Fig. \ref{pruning-dc40}. The training setting is kept the same for every model: learning rate $\tau=3\times 10^{-3},3\times 10^{-4},3\times 10^{-5},3\times 10^{-6}$ for 200, 200, 100 and 100 epochs, respectively, to guarantee the complete convergence of every competitor to ensure the fairness of comparison. For our method, the models are trained with C-SGD and trimmed. For Magnitude-~\cite{li2016pruning}, APoZ-~\cite{hu2016network} and Taylor-expansion-based~\cite{molchanov2016pruning}, the models are pruned by the different criteria and finetuned. The models labeled as Lasso are trained with group-Lasso Regularization for 600 epochs in advance, pruned, then finetuned for \textit{another} 600 epochs with the same learning rate schedule, such that the comparison is actually biased towards the Lasso method. The models are tested on the validation set every 10,000 iterations (12.8 epochs), and the collected results reveal the superiority of C-SGD in terms of the higher accuracy and also the better stability. Especially, though group-Lasso Regularization can indeed reduce the performance drop caused by pruning, it is outperformed by C-SGD by a large margin. Another interesting observation is that the models pruned by the importance metrics are unstable and trapped into a bad local minimum, which is consistent with~\cite{liu2018rethinking}, as the accuracy curves increase steeply in the beginning but slightly decline afterwards.

The above observations suggest it is better to train a redundant network and equivalently transform it into a narrower one than to finetune it after pruning. This is consistent with prior works~\cite{denton2014exploiting,hinton2015distilling} which pointed out that the redundancy in neural networks was necessary to overcome a highly non-convex optimization.

\subsection{Redundant Training \vs Normal Training}\label{sec-redundant-train}

In this subsection, we continue to verify the significance of training with manipulated redundant filters. However, we need to eliminate the effects of the well-trained base model, or we cannot tell whether the difference in the final accuracy is due to the redundancy during the training process or the powerful weights of the well-trained big model. Concretely, we train a narrow CNN with normal SGD and compare it with another model trained using C-SGD with the equivalent width \textit{from scratch}. To be specific, after random initialization, the latter produces some identical filters during C-SGD training, and when such filters are trimmed afterwards, it will have the same width as the former. For example, if a model has $2\times$ number of filters as the normal counterpart but every two filters are identical, they will end up with the same structure. In this way, if the redundant one outperforms the normal one, we can conclude that C-SGD can yield more powerful networks by exploiting the redundant filters.
\begin{table}[t]
	\caption{Top1 accuracy of scratch-trained DenseNet-40, VGG and ResNets using normal SGD or C-SGD with identical filters.}
	\label{exp-table-redundant-normal}
	\begin{center}
		\begin{small}
			\begin{tabular}{llcc}
				\toprule
				Dataset	&Model	   	& Normal SGD 	&C-SGD	\\
				\midrule
				CIFAR-10	&	DenseNet-3	&88.60	&\textbf{89.96}	\\
				CIFAR-10	&	DenseNet-6	&89.96	&\textbf{90.89}	\\
				CIFAR-10	&	VGG-1/4		&90.16	&\textbf{90.64}	\\
				CIFAR-10	&	VGG-1/2		&92.49	&\textbf{93.22}	\\
				CIFAR-10	&	ResNet-56-10-20-40	&91.78		&\textbf{92.81}	\\
				ImageNet	&	ResNet-50-30		&69.67		&\textbf{72.54}	\\
				\bottomrule
			\end{tabular}
		\end{small}
	\end{center}
\end{table}


On DenseNet-40, we evenly divide the 12 filters at each incremental convolutional layer into 3 clusters, use C-SGD to train the network from scratch, then trim it to obtain a model with 3 filters per incremental layer. \Ie, every 4 filters are growing centripetally. As a contrast, we train a DenseNet-40 with originally 3 filters per layer by normal SGD. Another group of experiments where each layer ends up with 6 filters are carried out similarly. After that, we experiment on VGG, slimming each layer to 1/4 and 1/2 of the original width, respectively. We also conduct experiments on ResNet-56 with target structure 10-20-40 and on ResNet-50 where every internal layer is reduced to 30\% of the original width.

It can be concluded from Table. \ref{exp-table-redundant-normal} that the redundant filters do help, compared to a normally trained counterpart with the equivalent width. This observation supports our intuition and assumption that the centripetally growing filters can enhance the model's representational capacity because though these filters are constrained, their corresponding input channels of the succeeding layers are still in full use and can grow without constraints in the parameter hyperspace (Fig. \ref{motivation-sketch}).

In other words, though C-SGD is originally designed for filter pruning on an off-the-shelf model, in some cases when a well-trained model is unavailable, we can still use C-SGD to train a wide redundant model from scratch and trim it into the desired structure. Though doing so delivers a lower accuracy than pruning a well-trained model (e.g., 92.81 \vs 93.62 on ResNet-56-10-20-40, as reported in Table. \ref{exp-table-redundant-normal} and \ref{exp-table-cifar}), we can still obtain a more powerful model than training from scratch using normal SGD.

\subsection{Global Slimming \vs Clipping Some Layers}\label{sect-exp-slim-clip}

In this subsection, we show that with the same target FLOPs, global ``slimming'' yields better results than simply ``clipping'' some of the layers. Concretely, we prune a ResNet-56 on CIFAR-10 to reach a comparable level of FLOPs as C-SGD-Res56-10-20-40 (Table. \ref{exp-table-cifar}). Instead of slimming every layer to 5/8 of the original width, we use C-SGD to prune the internal layers only, \ie, the first layers in each residual block. To realize 60\% FLOPs reduction, we slim such layers to 3/8 of the original width. We use $[x, y]$ to denote the structure of a ResNet stage where the first layer in every residual block has $x$ filters and the second has $y$. 

As shown in Table. \ref{exp-table-clipping}, clipping the internal layers delivers a significantly lower accuracy, which demonstrates the superiority of global slimming over simply clipping some layers in maintaining the accuracy, given a specific overall pruning ratio.

\begin{table}[t]
	\caption{Top1 accuracy of ResNet-56 pruned by global slimming or clipping the internal layers.}
	\label{exp-table-clipping}
	\begin{center}
		\begin{small}
			\begin{tabular}{llcc}
				\toprule
				&	Resulting width	   		& Top1 acc 	&FLOPs $\downarrow$	\\
				\midrule
				Global slimming		&	[10,10]-[20,20]-[40,40]	&\textbf{93.62}		&60.85	\\
				Clipping	&	[6,16]-[12,32]-[24,64]	&91.77				&61.76	\\
				\bottomrule
			\end{tabular}
		\end{small}
	\end{center}
\end{table}

\subsection{Scaling and Squeezing for More Powerful CNNs}\label{sect-scale-squeeze}
\begin{table*}
	\caption{Scaling and Squeezing on VGG and ResNet-50. Of note is that we calculate the required FLOPs of every model in the same manner as Luo \etal~\cite{luo2017thinet,DBLP:journals/pami/LuoZZXWL19}, such that the FLOPs are $2\times$ as those reported in other papers~\cite{he2016deep,li2016pruning}.}
	\label{exp-table-scale-prune}
	\begin{center}
		\resizebox{\textwidth}{!}{
			\begin{tabular}{lllccccc}
				\toprule
				Dataset		&	Model		&	Result 				& Top1	&FLOPs 		&	Layer Width\\
				\midrule
				CIFAR-10	&	VGG			&	Baseline			&	93.53	&	626M		&	64-128-256-512	\\
				CIFAR-10	&	VGG			&	$2\times$ scaled	&	93.69	&	2499M		&	128-256-512-1024	\\
				CIFAR-10	&	VGG			&	$2\times$ pruned	&	\textbf{93.97}	&	626M		&	64-128-256-512	\\
				\midrule
				ImageNet	&	ResNet-50	&	Baseline						&	75.33		&7.71B	&	64-[64-64-256]-[128-128-512]-[256-256-1024]-[512,512,2048]	\\
				ImageNet	&	ResNet-50	&	Global $1.25\times$				&	76.97		&11.98B	&	80-[80-80-320]-[160-160-640]-[320-320-1280]-[640,640,2560]	\\									
				ImageNet	&	ResNet-50	&	Global $1.25\times$ pruned		&	\textbf{76.23}		&7.71B	&	64-[64-64-256]-[128-128-512]-[256-256-1024]-[512,512,2048]\\
				ImageNet	&	ResNet-50	&	Bottleneck $2\times$			&	76.82		&13.05B	&	64-[64-128-256]-[128-256-512]-[256-512-1024]-[512,1024,2048]	\\
				ImageNet	&	ResNet-50	&	Bottleneck $2\times$ pruned		&	75.88		&7.71B	&	64-[64-64-256]-[128-128-512]-[256-256-1024]-[512,512,2048]\\
				\bottomrule
			\end{tabular}
		}
	\end{center}
\end{table*}

Scaling and Squeezing is a novel methodology to improve the performance of CNN based on C-SGD. The resulting model will deliver a higher level of accuracy with the same computational budgets as a normally trained counterpart. Concretely, we choose a mature CNN as the baseline, train a network with the same architecture but wider layers from scratch using regular SGD, and then use C-SGD to squeeze it into the original width.

On CIFAR-10, a $2\times$ scaled VGG is trained from scratch with normal SGD, \ie, every layer of the model is $2\times$ as wide as the normal VGG architecture. We then slim it down to the original structure by pruning half of the filters at each layer. On ImageNet, we train a $1.25\times$ scaled ResNet-50 from scratch, and slim it down to the original structure. Note that \textit{every} convolutional layer is widened to $1.25\times$ of its original width, including the pacesetters and followers, which are considered troublesome by the prior works. We then use C-SGD to prune every layer simultaneously. We also experiment with another model scaled differently, where only the bottleneck layers (\ie, the internal $3\times3$ convolutional layers in the residual blocks) are scaled by $2\times$.

As shown in Table. \ref{exp-table-scale-prune}, the pruned models consistently beat the counterparts trained with regular SGD by a clear margin. Intuitively, when the filters in each cluster are constrained to grow closer, the learned knowledge is gradually ``squeezed'' into the cluster center, \ie, the merged filters, such that the resulting model becomes more powerful than the normal counterpart. Interestingly, Global 1.25$\times$ pruned outperforms Bottleneck $2\times$ pruned (76.25 \vs 75.88 Top1 accuracy, 0.74 \vs 0.94 error increase), though Bottleneck $2\times$ requires more computations. It suggests that scaling and pruning globally, including those troublesome layers, yields better performance than only scaling and pruning the easy-to-prune layers. This observation again highlights the significance of C-SGD in solving the constrained filter pruning problem together with the results shown in Sect. \ref{sect-exp-slim-clip}.

\section{Discussions}

\subsection{C-SGD is Efficient}

The total time required for pruning is determined by the training (plus finetuning, if any) epochs, the training speed, the time consumed by the other algorithms (for those non-end-to-end methods), and the pruning granularity (\ie, the number of layers/filters to prune at a time). C-SGD is efficient because it is end-to-end, requires no finetuning, runs as fast as regular SGD and prune all the layers simultaneously. 

\textbf{C-SGD requires no finetuning after pruning.} As shown in Sect. \ref{sec-vs-zero-out}, during C-SGD training, the filters in each cluster can become \textit{infinitely close} to each other at a \textit{constant rate} with a constant learning rate. This property shows the superiority of identical-filter redundancy pattern over the small-norm pattern, as the latter cannot zero out the filters, but only reduce the magnitude of their parameters. As trimming the identical filters causes no performance drop, there is no need for a finetuning process, which is essential in many prior works~\cite{polyak2015channel,hu2016network,li2016pruning,alvarez2016learning,molchanov2016pruning,wen2016learning,abbasi2017structural,anwar2017structured,he2017channel,liu2017learning,luo2017thinet,he2018adc,yu2018nisp}.

\textbf{C-SGD allows one-step pruning on very deep CNNs.} The effectiveness and efficiency of C-SGD on very deep CNNs distinguish C-SGD from the layer-by-layer~\cite{polyak2015channel,hu2016network,alvarez2016learning,he2017channel,luo2017thinet,he2018adc} or filter-by-filter~\cite{molchanov2016pruning,abbasi2017structural} pruning methods. Many prior works choose to prune layer by layer because pruning too many layers at once may damage the network so severely that it cannot be finetuned to reach a satisfactory level of accuracy. In addition, the relative importance of filters is usually affected by the subsequent layers~\cite{yu2018nisp}, such that pruning several layers stacked together at once may lead to poor estimation of the importance of filters. In contrast, C-SGD can produce the desired redundancy patterns on all the layers simultaneously to prune them all at once. In practice, we observe no accuracy drop caused by the trimming step, even in very deep CNNs like ResNet-164 and DenseNet-121.

\textbf{C-SGD introduces negligible extra computational burdens.} We construct the averaging matrix $\bm{\Gamma}$ and decaying matrix $\bm{\Lambda}$ according to the clustering results $\mathcal{C}$ as two constants, and store them in the GPU memory. Compared to normal SGD, for each kernel tensor at each training iteration, the only extra computations introduced are two matrix multiplications (Eq. \ref{matrix-mul}), which consume minimal extra time and energy. In practice, the difference in the training speed between C-SGD and normal SGD is not observed. As a contrast, group-Lasso slows down the training significantly, as it requires costly square root operations.

\subsection{C-SGD is Robust to the Centripetal Strength}\label{sec-insensitive}
\begin{figure}
	\centering
	\subfloat[Value of $\chi$.]
	{
		\includegraphics[width=0.46\linewidth]{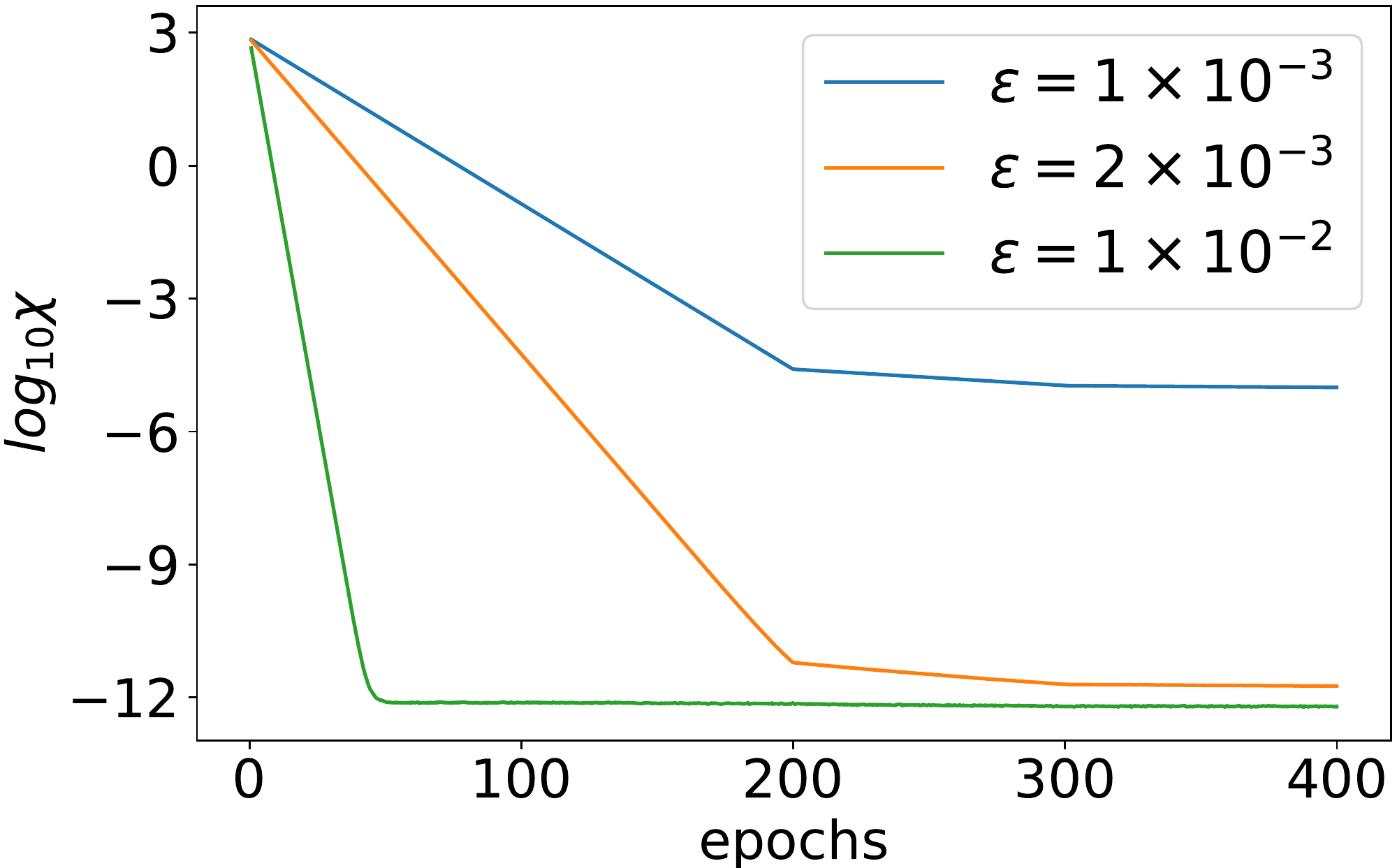}
		\label{hyper-parameter-curves-left}
	}
	\subfloat[Training loss.]
	{
		\includegraphics[width=0.46\linewidth]{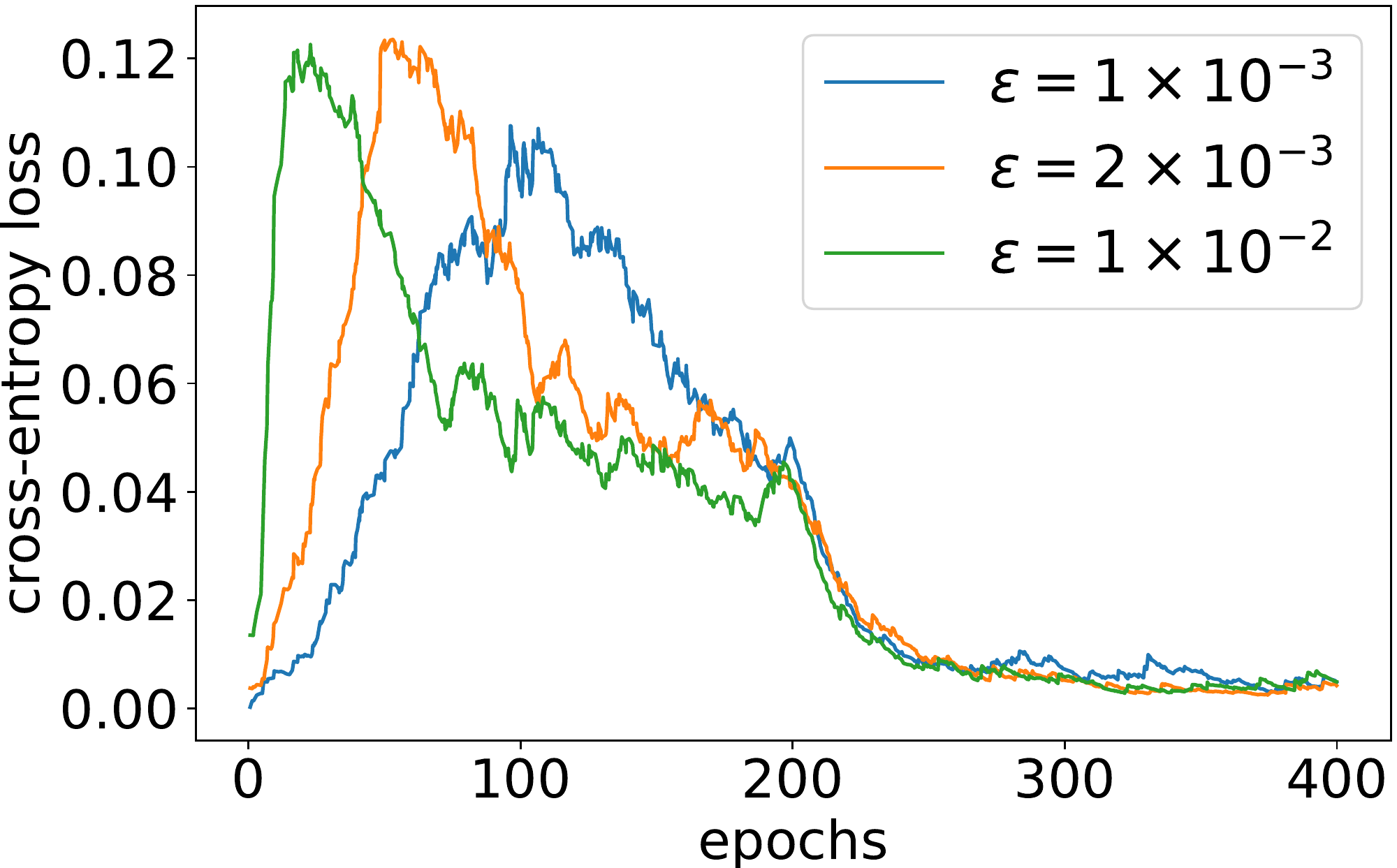}
		\label{hyper-parameter-curves-right}
	}
	\caption{Curves of the sum of squared kernel deviation $\chi$ (Eq. \ref{eq-def-chi}, note the logarithmic scale) and the training loss with different centripetal strength $\epsilon$. The learning rate is decayed at epoch 200 and 300, respectively.}
	\label{hyperparameter-curves}
\end{figure}

We perform a set of controlled experiments on ResNet-56 to study the effects of the centripetal strength $\epsilon$ by setting $\epsilon$ to $1\times10^{-3}$, $2\times10^{-3}$ and $1\times10^{-2}$, respectively. Fig. \ref{hyperparameter-curves} shows that C-SGD is robust to $\epsilon$, as the three models converge in a similar way. Intuitively, when we use C-SGD to produce the redundancy patterns on \textit{every} layer simultaneously, the network undergoes a period of progressive change, which leads to an increasing loss. When this kind of change becomes stable, \ie, when the filters in each cluster have become close enough, the loss starts to decrease. Obviously, with a larger $\epsilon$, the filters in each cluster grow centripetally at a faster rate, thus the change is finished earlier. 

\section{Conclusion}
We proposed to manipulate redundancy patterns by making some filters identical for pruning. By C-SGD, we have \textbf{1)} partly solved an open problem of constrained filter pruning on very deep CNNs with complicated architectures and shown the significance of such a success, \textbf{2)} beaten many recent competitors on several common benchmarks, \textbf{3)} presented empirical evidence for the assumption that redundancy can help the convergence of neural network training, which may encourage future studies, and \textbf{4)} proposed Scaling and Squeezing, a training methodology to boost the performance of CNNs.

\ifCLASSOPTIONcaptionsoff
  \newpage
\fi



%

\bibliographystyle{IEEEtran}      
\bibliography{csgdbib}   

%

%
%
%




\end{document}